\documentclass[letterpaper, 10 pt, conference]{ieeeconf}

\IEEEoverridecommandlockouts
\usepackage[english]{babel}
\usepackage[T1]{fontenc}

\usepackage[nolist]{acronym}
\usepackage[cmex10]{amsmath}
\usepackage{amssymb}
\usepackage{bm}
\usepackage{booktabs}
\usepackage{color}
\usepackage{float}
\usepackage{graphicx}
\usepackage{hyperref}
\usepackage{import}
\usepackage{mathtools}
\usepackage{multirow}
\usepackage{outline}
\usepackage{paralist}
\usepackage{siunitx}
\usepackage{stmaryrd}
\usepackage{systeme}
\usepackage{url}
\usepackage{tikz}
\usepackage{booktabs}

\usepackage{pgfplots}
\pgfplotsset{compat=1.3}
\usepackage{xstring}
\usepackage{subfigure}
\usepackage[overload]{empheq}
\usetikzlibrary{arrows.meta}
\usepackage{ulem}

\newtheorem{remark}{Remark}

\newcommand{\figref}[1]{Fig.~\ref{#1}}

\newcommand{\mnorm}[1]{\left\| #1 \right\|}

\newcommand{\R}{\mathbb{R}}

\newcommand{\x}{\mathbf{x}}
\newcommand{\bu}{\mathbf{u}}
\newcommand{\dt}{\text{d}t}
\newcommand{\f}{\mathbf{f}}
\newcommand{\g}{\mathbf{g}}
\newcommand{\zero}{\mathbf{0}}
\newcommand{\h}{\mathbf{h}}%

\newcommand{\p}{\mathbf{p}}
\newcommand{\q}{\mathbf{q}}
\newcommand{\bv}{\mathbf{v}}
\newcommand{\bl}{\mathbf{l}}

\newcommand{\br}{\mathbf{r}}

\newcommand{\dlambda}[1]{\bm \lambda_k^{\mathcal{#1}{\IfEq{#1}{O}{_i}{}}}}
\newcommand{\clambda}[1]{\bm \lambda_{\mathcal{#1}{\IfEq{#1}{O}{_i}{}}}}
\newcommand{\bb}[1]{\mathbf{b}_{\mathcal{#1}{\IfEq{#1}{O}{_i}{}}}}
\newcommand{\by}[1]{\mathbf{y}_k^{\mathcal{#1}{\IfEq{#1}{O}{_i}{}}}}
\newcommand{\bA}[1]{A_{\mathcal{#1}_{\IfEq{#1}{O}{_i}{}}}}
\newcommand{\btau}{\bm \tau}

\definecolor{my_blue}{rgb}{0, 0.4470, 0.7410}
\definecolor{my_yellow}{rgb}{0.9290, 0.6940, 0.1250}
\definecolor{my_purple}{rgb}{0.4940, 0.1840, 0.5560}
\definecolor{my_green}{rgb}{0.4660, 0.6740, 0.1880}
\definecolor{my_red}{rgb}{0.6350, 0.0780, 0.1840}
\definecolor{my_black}{rgb}{0.25, 0.25, 0.25}

\newcommand{\plotdis}[1]{
    \begin{tikzpicture}
    \pgfplotsset{
        width=0.5\linewidth, height=0.37\linewidth, 
        major tick length = 0.05cm, 
        grid style=dashed, label style={font=\scriptsize}, tick label style={font=\scriptsize},
        x label style={at={(axis description cs:0.5,-0.15)}},
        y label style={at={(axis description cs:-0.15,.5)}},
    }
    \begin{axis}[xlabel = {Time [s]}, ylabel = {$d_{\mathcal{O}_j}$ [m]}]
        \addplot[my_blue, line width=0.5pt] table[col sep=comma, x = time, y = /distances/data.0] {data/dis_#1.csv};
        \addplot[my_orange, line width=0.5pt] table[col sep=comma, x = time, y = /distances/data.1] {data/dis_#1.csv};
        \addplot[my_yellow, line width=0.5pt] table[col sep=comma, x = time, y = /distances/data.2] {data/dis_#1.csv};
        \IfEq{#1}{line}{}
        {
        \IfEq{#1}{l}{}{\addplot[my_purple, line width=0.5pt] table[col sep=comma, x = time, y = /distances/data.3] {data/dis_#1.csv};}
        }
        \addplot[my_black, line width=0.8pt, dashed] table[col sep=comma, x = time, y = /distances/header/stamp] {data/dis_#1.csv};

    \end{axis}
    \end{tikzpicture}
}

\newcommand{\plotcomputationtime}{
\begin{tikzpicture}
    \pgfplotsset{
        legend pos=north west,
        label style={font=\large},
    }
\begin{axis}[
        xlabel = {Number of obstacles},
	ylabel = {Computation time [ms]},
	enlargelimits=0.15,
	legend style={legend columns=1},
	ybar,
	bar width=7pt,
]
\addplot[fill=my_blue, my_blue] table[col sep=comma, x = num_obstalce, y = duality/average] {data/computation_time.csv};
\addplot[fill=my_yellow, my_yellow] table[col sep=comma, x = num_obstalce, y = duality_cbf/average] {data/computation_time.csv};

\legend{Exponential-DCBF Duality, Duality}
\end{axis}
\end{tikzpicture}
}

\title{\bf Walking in Narrow Spaces: Safety-critical Locomotion Control for Quadrupedal Robots with Duality-based Optimization}
\author{Qiayuan Liao, Zhongyu Li, Akshay Thirugnanam, Jun Zeng, and Koushil Sreenath
\thanks{This work is supported in part by NSF Grants CMMI-1931853 and CMMI-1944722.}
\thanks{All authors are with the University of California, Berkeley, USA. {\tt\small \{qiayuanl, zhongyu\_li, akshay\_t, zengjunsjtu, koushils\}@berkeley.edu}.}
}

\begin{document}
\maketitle

\begin{abstract}
This paper presents a safety-critical locomotion control framework for quadrupedal robots.
Our goal is to enable quadrupedal robots to safely navigate in cluttered environments.
To tackle this, we introduce exponential Discrete Control Barrier Functions~(exponential DCBFs) with duality-based obstacle avoidance constraints into a Nonlinear Model Predictive Control~(NMPC) with Whole-Body Control~(WBC) framework for quadrupedal locomotion control.
This enables us to use polytopes to describe the shapes of the robot and obstacles for collision avoidance while doing locomotion control of quadrupedal robots. 
Compared to most prior work, especially using CBFs, that utilize spherical and conservative approximation for obstacle avoidance, this work demonstrates a quadrupedal robot autonomously and safely navigating through very tight spaces in the real world. (Our open-source code is available at \url{https://github.com/HybridRobotics/quadruped_nmpc_dcbf_duality}, and the video is available at \url{https://youtu.be/p1gSQjwXm1Q}.)
\end{abstract}

\section{Introduction}

\begin{figure}[!t]
    \centering
    \resizebox{0.8\linewidth}{!}{
    \begin{tikzpicture}
      \node[] at (0,0) {\includegraphics[width=\linewidth]{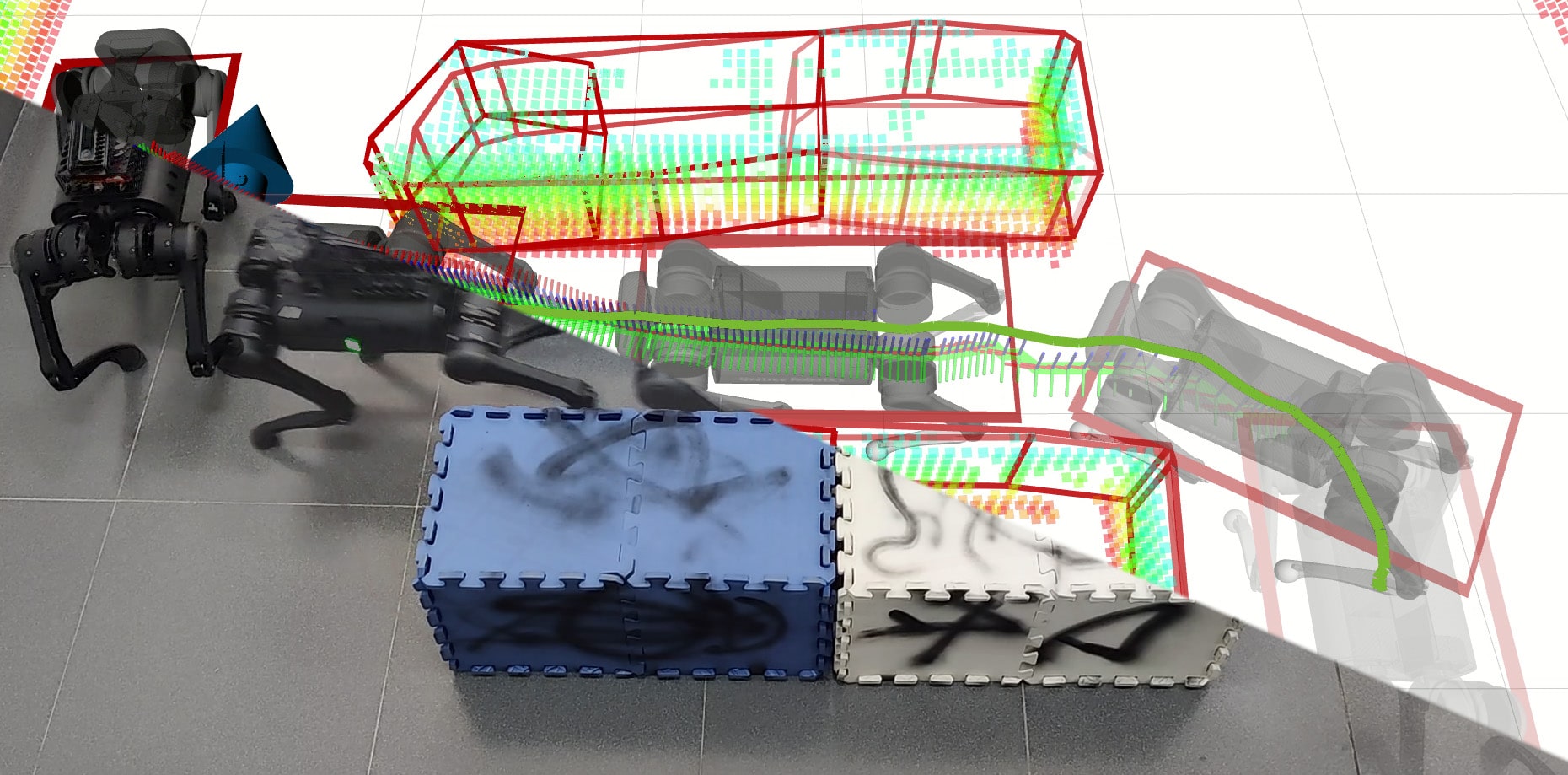}};
        \node[right] at (-0.5\linewidth, -0.22\linewidth) {\textbf{Real experiment}};
        \node[left] at (0.5\linewidth, 0.22\linewidth) {\textbf{Visualization}};
    \end{tikzpicture}
    }
    \caption{A quadrupedal robot whose width is $0.32$~m autonomously and safely navigates through a tight space that is only $0.5$-meter-wide using the proposed NMPC with exponential DCBF Duality framework in experiments.}
    \label{fig:cover}
    \vspace{-15pt}
\end{figure}

Legged robots have high maneuverability compared to wheeled robots, allowing them to traverse rough terrains and challenging environments and realize dynamic motions.
However, their highly nonlinear dynamics and hybrid modes make control of legged robots a challenging task.
Ensuring safe navigation through an obstacle-filled environment further exacerbates this problem~\cite{tranzatto2022cerberus}. 
Some prior work on legged locomotion achieves obstacle avoidance by decoupling the motion planning and the locomotion tasks, like in~\cite{buchanan2021perceptive, huang2021efficient, tonneau2018efficient}, where the motion planning problem is first solved for the legged robot's shape, and a control loop is used to track the planned path.
Since the planning is performed without the robot's dynamics, the trajectory is not necessarily dynamically feasible, which can violate safety constraints. 
Other works on obstacle avoidance for legged robots combine the planning and control tasks using optimizations such as model predictive control (MPC)~\cite{grandia2021multi}. 
However, most of these works enforce safety online by over-approximating the shape of the robots and obstacles, which results in conservative movements and can lead to deadlock~\cite{teng2021toward}.
To tackle this problem, this paper develops a nonlinear model predictive control (NMPC) formulation with discrete-time control barrier functions (DCBF) constraints for robot locomotion.
Unlike most of the prior work using CBFs~\cite{agrawal2017discrete,grandia2021multi}, the safety constraints are enforced by considering the polytopic shapes of the robot and its surrounding obstacles, allowing for tight obstacle avoidance motions in cluttered environments, as depicted in Fig.~\ref{fig:cover}. 
We validate our safety-critical locomotion algorithm with our proposed autonomy stack to achieve navigation tasks through tight spaces.

\subsection{Related Work}
Previous work tackling the navigation problem using legged robots can be broadly classified into three categories: (i) considering obstacles only in the motion planning layer without the robot's dynamics, (ii) including collision avoidance in optimal control but without considering the robot's finer shape, and (iii) safety-critical control for tight maneuvers.
\subsubsection{Collision-free Motion Planning}
Motion planning for legged robots has been an attractive topic and usually involves planning in the configuration space~\cite{tonneau2018efficient, buchanan2021perceptive}.
There are some approaches considering avoiding obstacles in confined spaces, as demonstrated in~\cite{schulman2014motion, kumagai2018efficient}.
Whole-body motion planning for the quadrupedal robot with signed distance field (SDF)~\cite{oleynikova2017voxblox} and elevation map are demonstrated in~\cite{buchanan2021perceptive} using a hierarchical motion planning framework.
However, most motion planning work for legged robots~\cite{dudzik2020robust,kim2020vision, buchanan2021perceptive} only considers planning in configuration space without the dynamics of the legged robot, which results in slow and statically stable gaits.
Additionally, the entire framework could fail due to collisions since the control layer doesn't consider the safety criteria reported in~\cite{buchanan2021perceptive}.
This motivates us to consider system dynamics while avoiding obstacles.

\subsubsection{Obstacle Avoidance in Trajectory Generation \& Control}
Obstacle avoidance with respect to system dynamics could be considered in both trajectory generation and control problems.
Some existing work considers the trajectory generation problem within an MPC formulation~\cite{gaertner2021collision, chiu2022collision}, where positivity of the SDF is regarded as an optimization constraint to ensure the robot's safety.
Other approaches usually consider obstacle avoidance constraints as enforcing distance functions to be positive in trajectory generation problems.
To make the distance functions differentiable, the robot and the obstacles are usually considered circular or spherical controlled regions for 2D or 3D navigation.
The optimization with distance function constraints is validated by experiments~\cite{gaertner2021collision, li2023autonomous,xiao2021robotic} on legged robots for trajectory generation.
This approach has been extended by using discrete-time control barrier functions~\cite{ames2019control} in quadruped stepping on discrete terrains~\cite{grandia2021multi} and humanoid navigation simulations~\cite{teng2021toward,li2022bridging,narkhede2022sequential}.
However, the main disadvantage of all the existing work above is that they over-approximate either the robot or the obstacles as circular or spherical objects in the trajectory generation or control problems. Although such a method could be efficient and sufficient for simple environments, it could be overly conservative and result in a deadlock maneuver in narrow spaces.
Additionally, although the SDF method tries not to over-approximate the robot by considering it as a combination of spherical objects, obstacles are still over-approximated by hyper-spheres.
This motivates us to develop autonomy algorithms to enable tight maneuvers for legged robot systems with less conservative safety criteria than the existing work. 

\subsubsection{Safety-Critical Tight Maneuvers}
To achieve non-conservative obstacle avoidance, polytopic obstacle avoidance is usually required, \textit{i.e.}, the robot and obstacles are considered as polytopes or combinations of them~\cite{zhang2021optimization}.
The main challenge of polytopic avoidance is that the distance function between polytopes is non-smooth.
Mixed-integer programming~\cite{grossmann2002review} could solve the problem but is only applicable for simple linear systems.
Dual analysis~\cite{zhang2021optimization} can reformulate the non-smooth polytopic constraints into smooth ones in the trajectory optimization problem for quadrupedal robots with its full-order dynamics offline~\cite{gilroy2021autonomous} or simplified dynamics online~\cite{yang2022collaborative}.
This dual analysis has also recently been synthesized by discrete-time control barrier functions (DCBFs)~\cite{zeng2021safety, zeng2021enhancing} to achieve real-time trajectory generation and control~\cite{thirugnanam2022safetycritical}.
As we will see here, we apply the dual optimization with a DCBF~\cite{thirugnanam2022safetycritical} to achieve safety-critical locomotion with less conservative maneuvers. 

\subsection{Contributions}
In this paper, we propose a robot autonomy stack that enables safety-critical locomotion in tight spaces with duality-based optimization. 
The primary contribution of this paper is the first introduction of duality-based Control Barrier Functions (CBFs) to legged locomotion control to ensure safety, \textit{e.g.}, obstacle avoidance, in the real world. 
The duality-based obstacle avoidance constraints allow us to describe the robot and its surrounding obstacles by polytopes. An ablation study shows that it provides a finer approximation of their shapes than the commonly-used spheres allowing the robot to traverse much tighter spaces with faster speed.
Combining with CBFs and model predictive control for the quadrupedal robot, we are able to realize non-conservative obstacle avoidance with smoother trajectory online.
Further, we develop an end-to-end navigation framework that combines robot perception feedback and the proposed safety-critical locomotion controller.
The proposed framework is deployed on a quadrupedal robot in the real world and enables the robot to autonomously and safely navigate through various narrow spaces in simulation and experiments.

\label{sec:Introduction}

\section{Background}
Before diving into the details of the proposed safety-critical locomotion framework, we first make a brief introduction of the background knowledge that is essential for the development of the following sections.

\subsection{Discrete-time Control Barrier Functions}
Consider a discrete-time dynamical system with states $\x_k \in \mathcal{X} \subset \mathbb{R}^n$ and inputs $\bu_k \in \mathcal{U}$ as
\begin{equation}
    \x_{k+1} = f_k^d(\x_k, \bu_k),
\end{equation}
where $\mathcal{U}$ is the admissible input set, which is compact.
In this paper, we consider a safe set of states $\mathcal{C}$, defined as the superlevel set of a continuously differentiable function $h: \mathcal{X} \subset \mathbb{R}^n \rightarrow \mathbb{R}$ by
\begin{equation}
    \begin{split}
        \mathcal{C} &= \{\mathbf{x} \in \mathbb{R}^n : h(\mathbf{x}) \geq 0 \}.
    \end{split}
    \label{eq:cbf-safeset}
\end{equation}
Throughout this paper, we refer to $\mathcal{C}$ as the safe set.
For robotics applications, the function $h$ can be the minimum distance function between the robot and the obstacle.
To guarantee safety, \textit{e.g.}, obstacle avoidance, we want to make $\mathcal{C}$ an invariant set, \textit{i.e.}, if the initial state lies in $\mathcal{C}$, then the entire evolution of the state trajectory should also lie in $\mathcal{C}$.

\begin{figure*}[!t]
    \centering
    \includegraphics[width=0.7\linewidth]{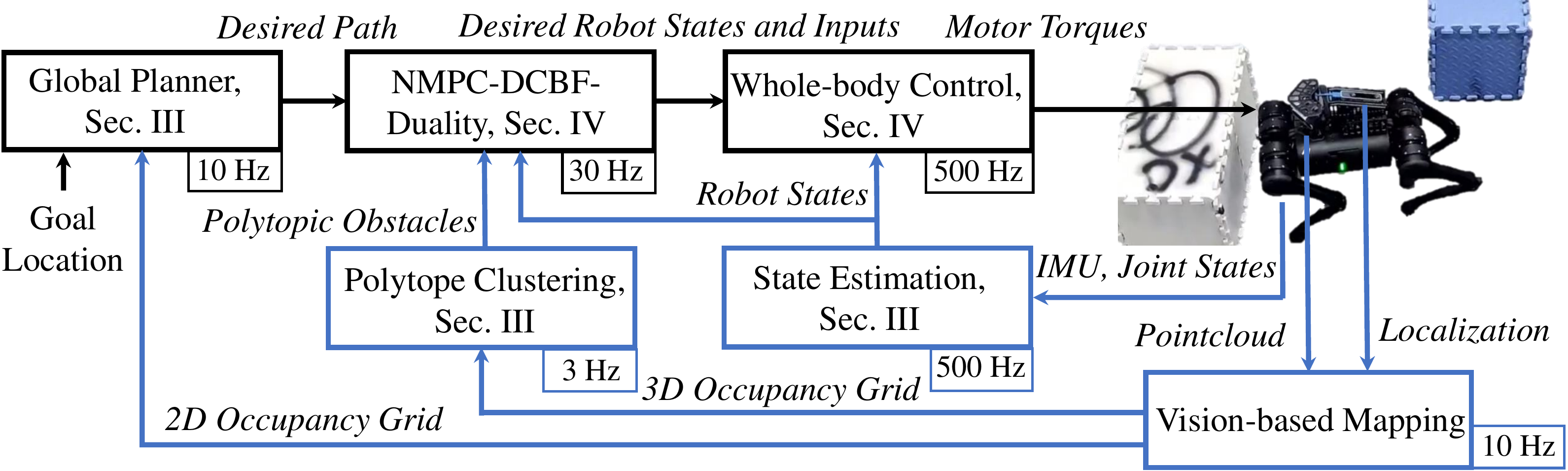}
    \vspace{-10pt}
    \caption{Overview of the proposed framework for safety-critical locomotion control of quadrupedal robots. The black blocks are related to planning and control, while the blue blocks are for state estimation and vision feedback. The robot uses two onboard depth cameras to perceive the world and leverages an additional tracking camera to localize itself. The sensed pointcloud is filtered and registered into a 3D occupancy grid (using Octomap~\cite{hornung2013octomap}) and clustered into polytopes. After being given a goal location, the global planner finds the desired path without considering the robot's shape from the robot's current position. The NMPC uses clustered polytopes for obstacle avoidance employing the exponential DCBF duality controller and tracking the desired global path while avoiding obstacles nearby. A higher-frequency WBC then computes the robot's motor torques based on the optimized results of the NMPC. Moreover, a state estimator using a Kalman Filter provides the currently estimated robot states to the NMPC and WBC.}
    \label{fig:system_dia}
    \vspace{-15pt}
\end{figure*}

The function $h$ becomes a discrete-time control barrier function if it satisfies the following relation $\forall \x_k \in \mathcal{X}$,
\begin{equation}
    h(\mathbf{x}_{k+1}) \geq \gamma(\x_k) h(\x_k), \quad 0 \leq \gamma(\x_k) \leq 1, \label{eq:dcbf}
\end{equation}
where $\x_{k+1} = f_k^d(\x_k, \bu_k)$ for some $\bu_k \in \mathcal{U}$, and $\gamma(\x_k)$ is the state-dependent decay rate~\cite{zeng2021enhancing}.
The DCBF constraint \eqref{eq:dcbf} enforces $h$ to decrease at most exponentially with the decay rate $\gamma(\x_k)$.

Given a choice of $\gamma(\x_k)$, we denote the set of feasible controls $\mathcal{K}(\x_k)$ as
\begin{equation}
    \begin{split}
        \mathcal{K}(\x_k) &\coloneqq \{\bu \in \mathcal{U} : h(f_k^d(\x_k, \bu_k)) \geq \gamma(\x_k)h(\x_k)\}.
    \end{split}
    \label{eq:cbf-safeinputset}
\end{equation}
Then if $\x_0 \in \mathcal{C}$ and $\bu_k \in \mathcal{K}(\x_k) \neq \emptyset, \;\forall k \in \mathbb{Z}^+$, then $\x_k \in \mathcal{C}$ for $\forall k \in \mathbb{Z}^+$, \textit{i.e.}, the resulting trajectory, is safe \cite{agrawal2017discrete}.

\subsection{Minimum Distance between Polytopes}
\subsubsection{Polytope representation}
We can describe the shape of the dynamic robot and the $i$-th static obstacle as convex polytopes in a $l$-dimensional space, which are defined using inequality constraints, respectively, as
\begin{subequations}
\begin{align}
    \mathcal{R}(\x_k) &\coloneqq \{ \mathbf{y} \in \mathbb{R}^l : \bA{R}(\x_k) \mathbf{y} \leq \bb{R}(\x_k) \}, \\
    \mathcal{O}_i &\coloneqq \{ \mathbf{y} \in \mathbb{R}^l : \bA{O} \mathbf{y} \leq \bb{O}\},
\end{align}
\end{subequations}
where $\mathcal{R}$ and $\mathcal{O}_i$ represent the robot and the $i$-th obstacle respectively, $\bA{R}(\x_k) \in \mathbb{R}^{s^{\mathcal{R}} \times l}, \bb{R}(\x_k) \in \mathbb{R}^{s^{\mathcal{R}}}$ define the robot, and $\bA{O} \in \mathbb{R}^{s^{\mathcal{O}_i} \times l}$ and $\bb{O} \in \mathbb{R}^{s^{\mathcal{O}_i}}$ define the $i$-th obstacle.
The symbols $s^{\mathcal{R}}$ and $s^{\mathcal{O}_i}$ represent the number of facets of the polytopic sets for the robot and $i$-th obstacle, respectively.
Note that the pose of the robot polytope $\mathcal{R}$ depends on the system state.

\subsubsection{Primal Problem}
The square of the minimum distance between $\mathcal{R}(\x_k)$ and $\mathcal{O}_i$, denoted by $d_i(\x_k)$, can be computed using a QP as follows:
\begin{subequations}
\begin{align}
d_i(\x_k)       &=  \min_{\by{R}, \by{O}} \mnorm{\by{R} - \by{O}}^2, \\
\textrm{s.t.} \quad & \bA{R}(\x_k) \by{R} \leq \bb{R}(\x_k), \ \bA{O} \by{O} \leq \bb{O}.
\end{align}
\label{eq:primal_problem}%
\end{subequations}
We directly want to use this as the DCBF function, \textit{i.e.},  $h_i(\x_k) = d_i(\x_k)$, and enforce $h_i(\x_k)\ge 0$.
However, the above optimization problem can only be solved numerically and not analytically.
This results in the difficulty of adding the minimum distance function in a DCBF constraint to other optimization problems because it is implicitly defined by \eqref{eq:primal_problem} and non-differentiable.

Moreover, the primal problem is a minimization problem and is not suitable for working with the DCBF constraint \eqref{eq:dcbf} \cite{thirugnanam2022safetycritical}.
Instead, we convert the minimization problem \eqref{eq:primal_problem} into a maximization problem using the principle of duality.

\subsubsection{Duality problem}
The dual problem of \eqref{eq:primal_problem} is given by
\begin{subequations}
\begin{align}
d_i(\x_k) &=\max_{\dlambda{R}, \dlambda{O}} -(\dlambda{R})^T\bb{R}(\x_k)-(\dlambda{O})^T\bb{O} \\
\textrm{s.t.} \quad & \bA{R}^T(\x_k) \dlambda{R} + \bA{O}^T \dlambda{O} = 0 , \\
& \mnorm{\bA{O}^T \dlambda{O}}_2 \leq 1, \ \dlambda{R} \geq 0, \ \dlambda{O} \geq 0, \label{eq:dual_problem_duality_norm}
\end{align}
\label{eq:dual_problem}%
\end{subequations}
where $\dlambda{R}$ and $\dlambda{O}$ are the dual variables corresponding to the primal problem.

Since \eqref{eq:dual_problem} is a SOCP that is slower to compute, we use another dual formulation \cite{thirugnanam2022safetycritical} for the distance computation, which results in a QP, and is given by
\begin{subequations}
\begin{align}
d_i(\x_k) &=  \max_{\dlambda{R}, \dlambda{O}}  -\frac{1}{4}(\dlambda{O})^T\bA{O}\bA{O}^T\dlambda{O}  \nonumber \\
                            & \quad -(\dlambda{R})^T\bb{R}(\x_k)-(\dlambda{O})^T\bb{O} \\
\textrm{s.t.} \quad         & \bA{R}^T(\x_k) \dlambda{R} + \bA{O}^T \dlambda{O} = 0 , \\
                            & \dlambda{R} \geq 0, \dlambda{O} \geq 0.
\end{align}
\label{eq:distance_qp}%
\end{subequations}
The QP \eqref{eq:distance_qp} results in the same optimal value $d_i(\x_k)$ as \eqref{eq:dual_problem}, but the optimal solution of \eqref{eq:distance_qp}, $\dlambda{R}$ and $\dlambda{O}$, is a scaled value of that of \eqref{eq:dual_problem}.

\subsection{Exponential DCBF Duality Constraints}
To reduce the complexity of the DCBF constraints when used in MPC, we make use of exponential DCBF constraints, where instead of enforcing \eqref{eq:dcbf} we enforce the following constraint,
\begin{equation}
\label{eq:exponential-dcbf}
h_i(\x_k) \geq (\Pi_{n=0}^{k} \gamma_{n}) h_i(\x_0),
\end{equation}
where $k$ represents the time index in the MPC and $\x_0$ is the current state.
Note that \eqref{eq:exponential-dcbf} is obtained by rolling out time in the DCBF constraint \eqref{eq:dcbf}.
It can be shown that the exponential DCBF constraint \eqref{eq:exponential-dcbf} is equivalent to the following set of exponential DCBF duality constraints \cite{thirugnanam2022safetycritical},
\begin{subequations}
    \begin{align}
        -(\dlambda{R})^T\bb{R}(\x_k) -(\dlambda{O})^T\bb{O} &\geq (\prod_{n=0}^k \gamma_n ) d_i(0) \label{eq:exp_dcbf_duality_lag} \\
        \bA{R}^T(\x_k) \dlambda{R} + \bA{O}^T \dlambda{O} &= 0, \label{eq:exp_dcbf_duality_vector}\\
        \mnorm{\bA{O}^T \dlambda{O}}_2 &\leq 1, \label{eq:exp_dcbf_duality_norm}\\
        \dlambda{R} \geq 0, \dlambda{O} &\geq 0, \label{eq:exp_dcbf_duality_positive}
    \end{align}
    \label{eq:exp_dcbf_duality}%
\end{subequations}
\noindent where $d_i(0) = d_i(\x_0)$ is the square of the minimum distance between the robot and the $i$-th obstacle at the current state $\x_0$, calculated using the distance QP \eqref{eq:distance_qp}.
As we will see, the above exponential DCBF duality constraints \eqref{eq:exp_dcbf_duality} will be incorporated into an NMPC to guarantee safety for the locomotion control of quadrupedal robots. 

\label{sec:background}

\section{Overview of Proposed Framework}

Having developed the math background, we next present an overview of our entire framework as illustrated in \figref{fig:system_dia}.

The robot is equipped with depth cameras and a tracking camera. The environment is perceived via the depth camera by a pointcloud, and the tracking camera estimates the odometry of the robot through Visual-Inertial Odometry.
With estimated robot location, the pointcloud is filtered and registered in a 3D occupancy grid by Octomap \cite{hornung2013octomap} at 10Hz.
To describe the shape of obstacles nearby, the entire registered pointcloud is clustered into multiple small polytopes according to Euclidean distance and the quickhull algthom \cite{barber1996quickhull}. 
To be more computationally efficient, we only consider clustering the voxels near the robot, \textit{i.e.}, in a local map, and such a polytope clustering runs at $3$ Hz. 
The clustered 3D polytopes are projected to the ground to find the 2D convex polytope representation of the obstacles.
The Octomap also generates a projection to the ground to obtain a 2D occupancy grid map for global planning.

The global planner generates a path at a low frequency using Dijkstra's algorithm without considering the robot's shape.
The path of waypoints for the robot is first generated, and the yaw angles are set to be corresponding to forward orientations along the path. 
The path is converted to a trajectory according to a preset velocity and nominal joint angle and then tracked by the following NMPC controller with an exponential DCBF with duality to consider avoidance of the bounding polytopes of obstacles while controlling the robot's locomotion.

The NMPC controller uses a centroidal dynamics model of the quadrupedal robot~\cite{orin2013centroidal} with the system states as the robot's base pose, base momentum, and joint positions, and the system inputs as the ground contact forces and joint velocities.
To avoid collisions with local obstacles, we add the exponential DCBF duality constraints \eqref{eq:exp_dcbf_duality} for polytope obstacle avoidance.
To ensure the robot's movement, the NMPC evaluates optimized system states and inputs, combined with other constraints, such as friction cone constraints.
A Whole-body Controller (WBC)~\cite{bellicoso2016perception} follows the MPC to compute the robot's optimal generalized acceleration, contact forces (ground reaction forces), and joint torques according to the optimized states and inputs from the NMPC.
The torque computed by WBC is set as a feed-forward term and is sent to the robot's motor controller.
Combined with joint-level PD commands, this could reduce the shock during foot contact and improve tracking performance.

Running in the same loop with WBC, a Kalman Filter estimates the robot's base position and velocity from the onboard measurements of base orientation, base acceleration, and feet positions obtained by measured joint positions and the robot's forward kinematics.  

\label{sec:framework}

\section{NMPC DCBF Dual Formulation}
In this section, we briefly describe the formulation of NMPC with the exponential DCBF with duality for obstacle avoidance and then present some implementation details.

\subsection{NMPC for Quadrupedal Locomotion} \label{subsec:baseline_nmpc}
Consider a Nonlinear Model Predictive Control formulation with horizon $N$ as
\begin{subequations}
\begin{align}
\min_{\{\x_k, \bu_k\}} \quad &\sum_{k=0}^{N-1} l_k(\x_k, \bu_k), && \label{eq:cost} \\
\textrm{s.t.} \quad     & \x_0 = \x(0), && \label{eq:inital_state} \\
                        & \x_{k+1} = f_k^d(\x_k, \bu_k), && k=0,...,N-1 \label{eq:dynamics}, \\
                        & \g_k(\x_k, \bu_k) = \zero, && k=0,...,N \label{eq:eq_constraint}, \\
                        & \h_k(\x_k, \bu_k) \geq \zero && k=0,...,N \label{eq:ineq_constraint},
\end{align}
\label{eq:ocp}%
\end{subequations}
where $\x_k$ is the state and $\bu_k$ is the input at time $k$, $\x(0)$ is the current state, $l_k$ is a time-varying stage cost.
We want to find the control input that minimizes the total cost subject to the initial state $\x(0)$, system dynamics $f_k^d(\x_k, \bu_k)$, and the general equality $\g_k(\x_k, \bu_k)$ and inequality $\h_k(\x_k, \bu_k)$ constraints.

\subsubsection{System Dynamics}
We use a previous NMPC formulation using centroidal dynamics of a quadruped described in \cite{sleiman2021unified}, where the system states $ \x \in \R^{24} $ and inputs $\bu \in \R^{24} $ are defined as
\begin{equation}
    \x= [\h_{com}^T, \q_b^T, \q_j^T]^T, \quad \bu = [\f_c^T, \bv_j^T]^T
\end{equation}
where $\q=[\q_b^T, \q_j^T]^T$ is the generalized coordinate, and the ZYX-Euler angle parameterization is assumed to represent the robot's torso’s orientation.
$\h_{com} = [\p_{com}^T, \bl_{com}^T]^T\in \R^6$ is the collection of the normalized centroidal momentum, $\f_c \in \R^{12}$ consists of contact forces at four contact points, i.e., four ground reaction force of foot.
$\q_j$ and $\bv_j$ are the joint positions and velocities.
The continuous-time system flow map is given by
\begin{equation}
    \frac{\textrm{d}}{\dt}
    \begin{bmatrix}
        \p_{com} \\ \bl_{com} \\ \q_b \\ \q_j
    \end{bmatrix} = 
    \begin{bmatrix}
       \frac{1}{m} \sum\limits_{i=1}^{4} \f_{c_i} + \g \\
       \frac{1}{m} \sum\limits_{i=1}^{4} \br_{com,c_i} \times \f_{c_i}  \\
       A_b^{-1} (\h_{com} - M_j \bv_j) \\
       \bv_j
    \end{bmatrix}
    \label{eq:continuous-time-flow-map}
\end{equation}
where $m$ is the robot total mass, $\br_{com,\mathrm{c_i}}$ is the position of the $i$-th foot w.r.t to the center of mass, while $A(q) = [A_b(q)\ A_j(q)] \in \R^{6 \times 18}$ is the centroidal momentum matrix which maps generalized velocities to centroidal momentum.
Readers can refer to \cite{orin2013centroidal} for more details.
The dynamics \eqref{eq:continuous-time-flow-map} can be discretized and expressed in the discrete-time form \eqref{eq:dynamics}.

\subsubsection{Cost}
The cost \eqref{eq:cost} is a quadratic tracking cost to follow a given full system state trajectory, including base pose (and/or its twist), normalized momentum, and nominal joint positions.

\subsubsection{Constraints}
The gait (periodic contact sequence) is predefined, so we can formulate the constraints to ensure that stance legs remain on the same footholds and the swing legs follow predefined curves for the feet' heights with zero contact force.
A friction cone constraint of each stance leg is also added to avoid slipping. For more details about constraints, please see \cite{sleiman2021unified}.

\subsection{Exponential DCBF Duality Constraint}
The NMPC shown in Sec.~\ref{subsec:baseline_nmpc} has not considered obstacle avoidance.
To achieve safety-critical locomotion control with obstacle avoidance, we add exponential DCBF duality constraint \eqref{eq:exp_dcbf_duality} to the NMPC \eqref{eq:ocp} by including \eqref{eq:exp_dcbf_duality_lag}, \eqref{eq:exp_dcbf_duality_norm}, \eqref{eq:exp_dcbf_duality_positive} as inequality constraints $\h_k(\x_k, \bu_k)$, and \eqref{eq:exp_dcbf_duality_vector} as equality constraints $\g_k(\x_k, \bu_k)$.

\subsubsection{Signed Distance}
In order to let the solver know how "intrusive" the robot is with the obstacles, we use signed distance duality formulation \cite{zhang2021optimization} by replacing the inequality norm constraint \eqref{eq:exp_dcbf_duality_norm} with an equality constraint
\begin{equation}
    \mnorm{\bA{O}^T \dlambda{O}}_2 = 1. \label{eq:singed_duality}
\end{equation}
Although the convex constraint \eqref{eq:exp_dcbf_duality_norm} is then turned into a non-convex equality constraint, \eqref{eq:singed_duality} is an equality constraint and can be handled through a projection method, and the solving time can be unaffected. 
The advantage of using such a signed distance formulation is that, even if it collides with the obstacle in the real world, \textit{e.g.}, due to external perturbations, the robot will move away from the obstacle. 

\subsubsection{Margin} The CBFs \eqref{eq:exp_dcbf_duality_lag} are modified with two constant small margins $\alpha, \beta \geq 0$ as
\begin{subequations}
\begin{align*}
    &-(\dlambda{R})^T\bb{R}(\x_k) -(\dlambda{O})^T\bb{O} \geq (\prod_{n=0}^k \gamma_n ) \tilde{d}_i(\x_0)  + \alpha, \nonumber
\end{align*}
\end{subequations}
where $\alpha$ is the minimum distance margin between the polytopes of the robot and obstacles.
$\tilde{d}_i(\x_0)$ is given by
\begin{equation}
\tilde{d}_i(\x_0) = \max\{d_i(\x_0) - \beta, 0\},
\end{equation}
where $\beta$ is a margin to prevent the robot from adopting a conservative strategy, \textit{e.g.}, taking a large detour or stopping in front of the obstacle at the earlier stage of the NMPC. 

\subsection{Optimization Setup}
We add the dual variables $\dlambda{R}$ and $\dlambda{O}$ into the system inputs of the optimal control problem \eqref{eq:ocp}.
To solve this, a multiple shooting method is leveraged to transcribe the optimal control problem to a nonlinear program (NLP) problem, and the NLP is solved using Sequential Quadratic Programming (SQP) where the QP subproblem is solved using HPIPM \cite{frison2020hpipm}.
For more details regarding the solving process and algorithm, we refer the reader to \cite{thirugnanam2022safetycritical}.
The numerical optimization is formulated in OCS2~\cite{ocs2}.
In the scenario where four obstacles (each with 15 vertices) are presented, the optimization can be solved in around 25 ms.

At each NMPC iteration, the initial guess of dual variables over the MPC horizon $\tilde{\bm{\lambda}}^{\mathcal{R}}_{k,MPC}$ and $\tilde{\bm{\lambda}}^{\mathcal{O}_i}_{k,MPC}$ is provided. 
We first solve the QP formulated in~\eqref{eq:distance_qp} using initial state of the NMPC $\x_0$  and set the initial guess based on its solution as:
\begin{align}
    \tilde{\bm{\lambda}}^{\mathcal{R}}_{k,MPC} &= \frac{{\bm{\lambda}}^{\mathcal{R}^*}_{k,QP}}{2d_i(\x_0)}, \ \tilde{\bm{\lambda}}^{\mathcal{O}_i}_{k,MPC} = \frac{{\bm{\lambda}}^{\mathcal{O}^*_i}_{k,QP}}{2d_i(\x_0)}. 
    \label{eq:inital_guess}
\end{align}
These initial guess remains constant over the MPC horizon.  
We empirically found that using such an initial guess, the computing time can be significantly improved.

\subsection{Whole-Body Control (WBC)}
After solving the above-mentioned NMPC with exponential DCBF duality, we can obtain the optimized position and/or velocity profiles of the base, joints, and contact force from the optimizer.
These are then utilized in a hierarchical optimization whole body controller \cite{bellicoso2016perception}, which computes the torque for each joint according to the optimized result from the NMPC in a prioritized way.
The decision variable of WBC is: $\x_{wbc} = [\ddot \q^T, \f_{c_i}^T, \btau^T]^T$, where $\ddot\q$ is the generalized acceleration, and $\btau$ is the torque of all actuated joints.

With multiple tasks (equality and inequality constraints) defined, the WBC solves the QP problem in the null space of the higher priority tasks' linear constraints and tries to minimize the slack variables of the inequality constraints. This approach can consider the full nonlinear rigid body dynamics and ensure strict task priority~\cite{bellicoso2016perception}. 

We implement the WBC using the Pinocchio rigid body library~\cite{carpentier2019pinocchio} and qpOASES QP solver~\cite{ferreau2014qpoases}. We also run state estimation in this same control loop.

\label{sec:formulation}

\section{Results}
\begin{figure}[t]
    \centering
        \begin{tikzpicture}
        \node at (-0.48\linewidth,0) {\includegraphics[width=0.24\linewidth]{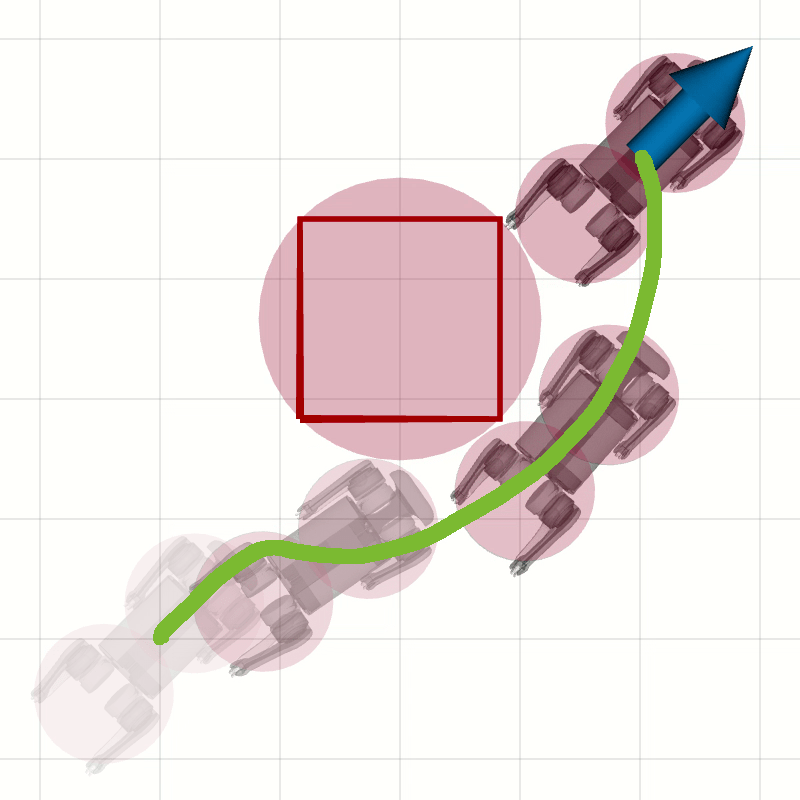}} node[xshift=-0.44\linewidth, yshift=-0.1\linewidth, scale=0.7]{(a)};
        \node at (-0.24\linewidth,0) {\includegraphics[width=0.24\linewidth]{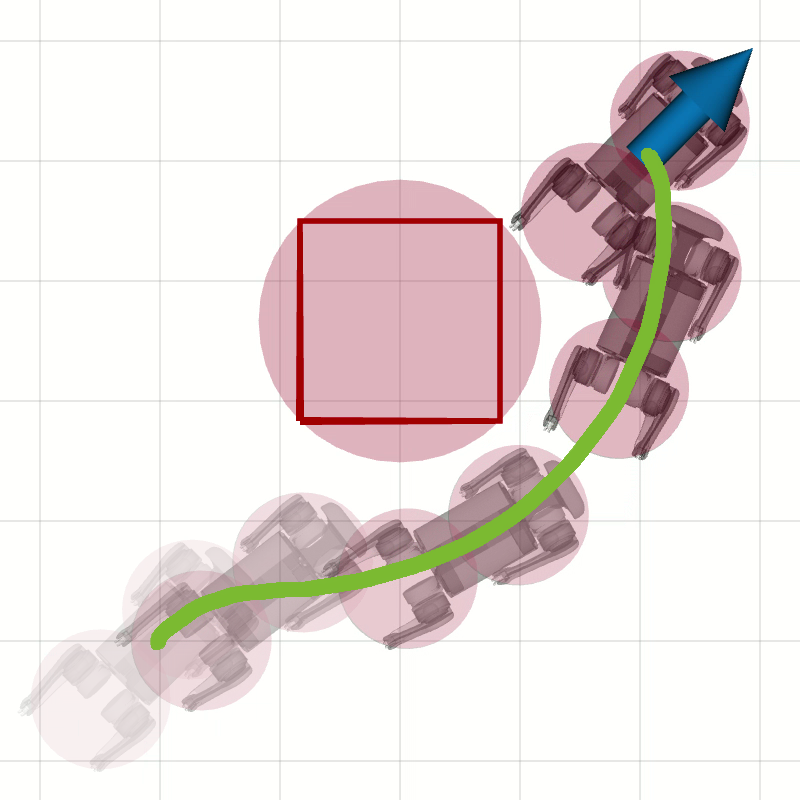}} node[xshift=-0.2\linewidth, yshift=-0.1\linewidth, scale=0.7]{(b)};
        \node at (0,0) {\includegraphics[width=0.24\linewidth]{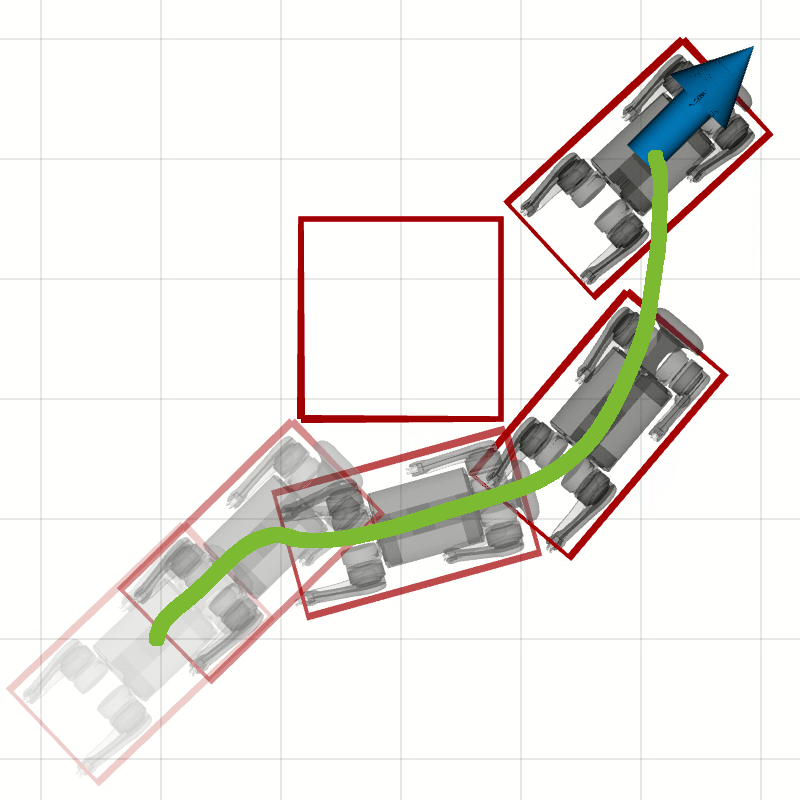}} node[xshift=0.04\linewidth, yshift=-0.1\linewidth, scale=0.7]{(c)};
        \node at (0.24\linewidth,0) {\includegraphics[width=0.24\linewidth]{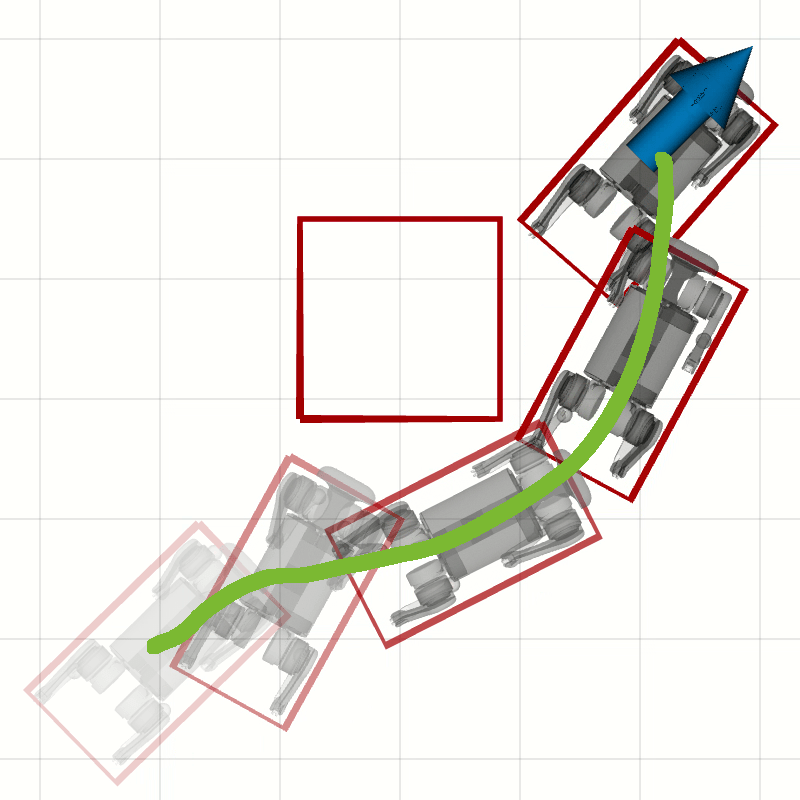}} node[xshift=0.28\linewidth, yshift=-0.1\linewidth, scale=0.7]{(d)};
        \end{tikzpicture}
    \caption{Snapshots for four different constraint formulations in simulation: (a) Euclidean Distance Constraint, (b) Euclidean Distance Exponential DCBF, (c) Duality Constraint, (d) Exponential DCBF Duality (ours). Lighter snapshots are earlier in time. Overall, the robot paths (green) using CBFs (b)(d) are smoother than the ones without CBF in (a)(c). By considering the obstacle's shape by duality formulation in (c)(d), the robot is able to avoid the obstacle less conservatively compared to the ones using Euclidean distance constraints in (a)(b).} 
    \label{fig:simple}
    \vspace{-15pt}
\end{figure}

After introducing the entire formulation for the proposed safety-critical locomotion controller for quadrupeds, we now move on to deploy the framework on a quadrupedal robot A1, which has 12 actuators and is \SI{0.3}{\m} wide and \SI{0.6}{\m} long. 
We validate our proposed algorithms in both simulation and experiments, and the results are recorded in the accompanying video. In the simulation, we first demonstrate the concept of NMPC with exponential DCBF duality constraints by avoiding a single obstacle using four different controllers. In order to highlight the advantages of the polytopic approximation, we further benchmark the proposed controller with a baseline in environments with random obstacles.

\subsection{Single Obstacle Simulation}~\label{subsec:single_sim}
We evaluate the proposed method in simulation using the robot dynamics in Gazebo.
To better demonstrate the effect of the duality and exponential DCBF constraints, we test four controllers in simulation as shown in \figref{fig:simple}.
The robot is commanded to a goal while avoiding a square obstacle in the middle between the starting and ending points.
There are four controllers which use different formulations for obstacle avoidance: 1) Euclidean Distance Constraint (\figref{fig:simple}a) where the robot and obstacle are approximated by circles, and then their separations ($l_2$ distances) are constrained to be positive over the entire MPC horizon. 
2) Euclidean Distance Exponential DCBF (\figref{fig:simple}b) where the $l_2$ distances between robot and obstacles are used as the function $h$ in exponential DCBF defined in~\eqref{eq:exponential-dcbf}. 
3) Duality Constraint (\figref{fig:simple}c) where the robot and obstacles are bounded by polytopes which are constrained to be separated by duality-based optimization~\cite{zhang2021optimization} without CBF, and 4) the proposed exponential DCBF Duality (\figref{fig:simple}d) where both polytopes and CBFs are used to avoid collisions.   

By comparing \figref{fig:simple}(a, b), which uses Euclidean distance that can only use circular shape approximation, and \figref{fig:simple}(c, d), which uses duality-based constraint that supports finer shape approximation by polytopes, we find that the robot has less detour in \figref{fig:simple}(c, d) than in \figref{fig:simple}(a, b). 
Such a property allows the robot to maneuver in a tighter space. 
Furthermore, by introducing the exponential DCBF, the robot is able to react to the obstacles earlier, such as \figref{fig:simple}b compared to \figref{fig:simple}a, and \figref{fig:simple}d compared to \figref{fig:simple}c. 
Therefore, the robot shows a smoother trajectory in \figref{fig:simple}(b,d) and can avoid the obstacle without having to make a sudden change in heading when the robot is close to the obstacle in \figref{fig:simple}(a,c).

\subsection{Advantages of Travelling in Narrow Spaces}~\label{subsec:narrow_sim}

\begin{figure}[!t]
    \centering
    \resizebox{0.9\linewidth}{!}{
    \begin{tikzpicture}
        \node[xshift=-0.25\linewidth] {\includegraphics[width=0.5\linewidth]{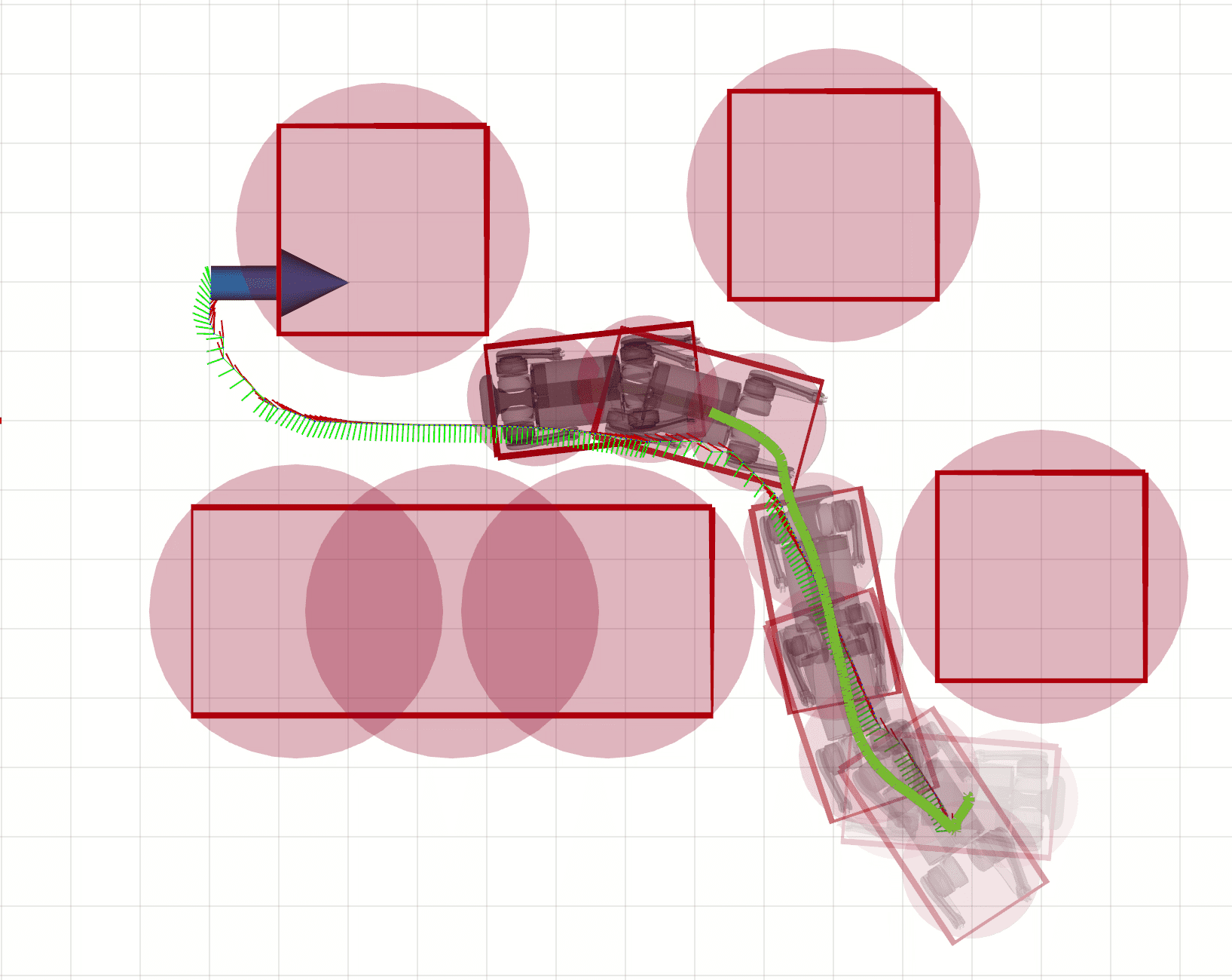}}  node[xshift=-0.25\linewidth] at (-0.2\linewidth, -0.15\linewidth) {(a)};
        \node[xshift=0.25\linewidth] {\includegraphics[width=0.5\linewidth]{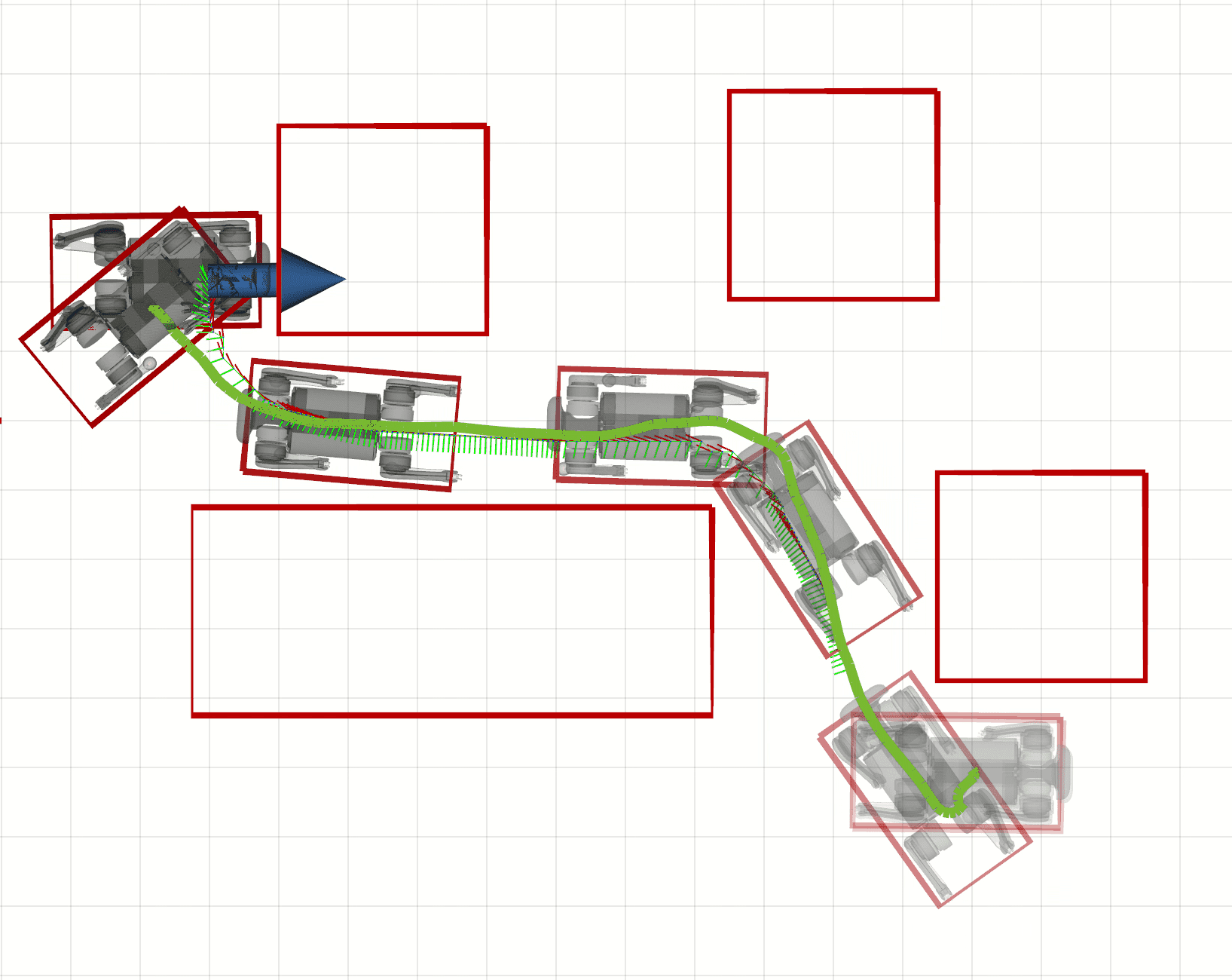}}  node[xshift=0.25\linewidth] at (-0.2\linewidth, -0.15\linewidth) {(b)};
        \draw[xshift=-0.25\linewidth, -{Stealth[scale=1.0]}] (0.02\linewidth, 0.1\linewidth) node[above,scale=1.0]{Get stuck!} -- (-0.03\linewidth, 0.05\linewidth);
    \end{tikzpicture}
    }
    \resizebox{0.9\linewidth}{!}{
    \begin{tikzpicture}
        \node[xshift=-0.25\linewidth] {\includegraphics[width=0.5\linewidth]{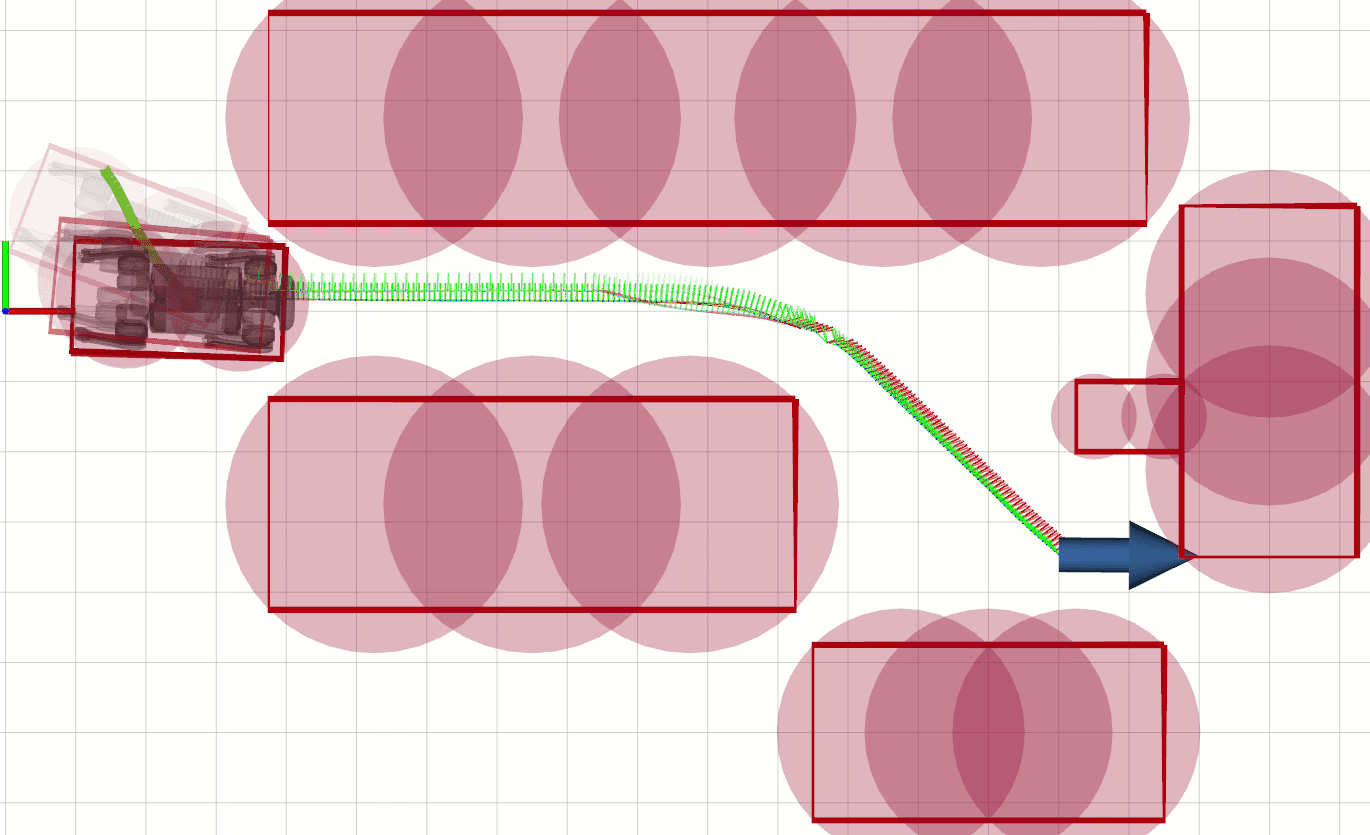}}  node[xshift=-0.25\linewidth] at (-0.2\linewidth, -0.15\linewidth) {(c)};
        \node[xshift=0.25\linewidth] {\includegraphics[width=0.5\linewidth]{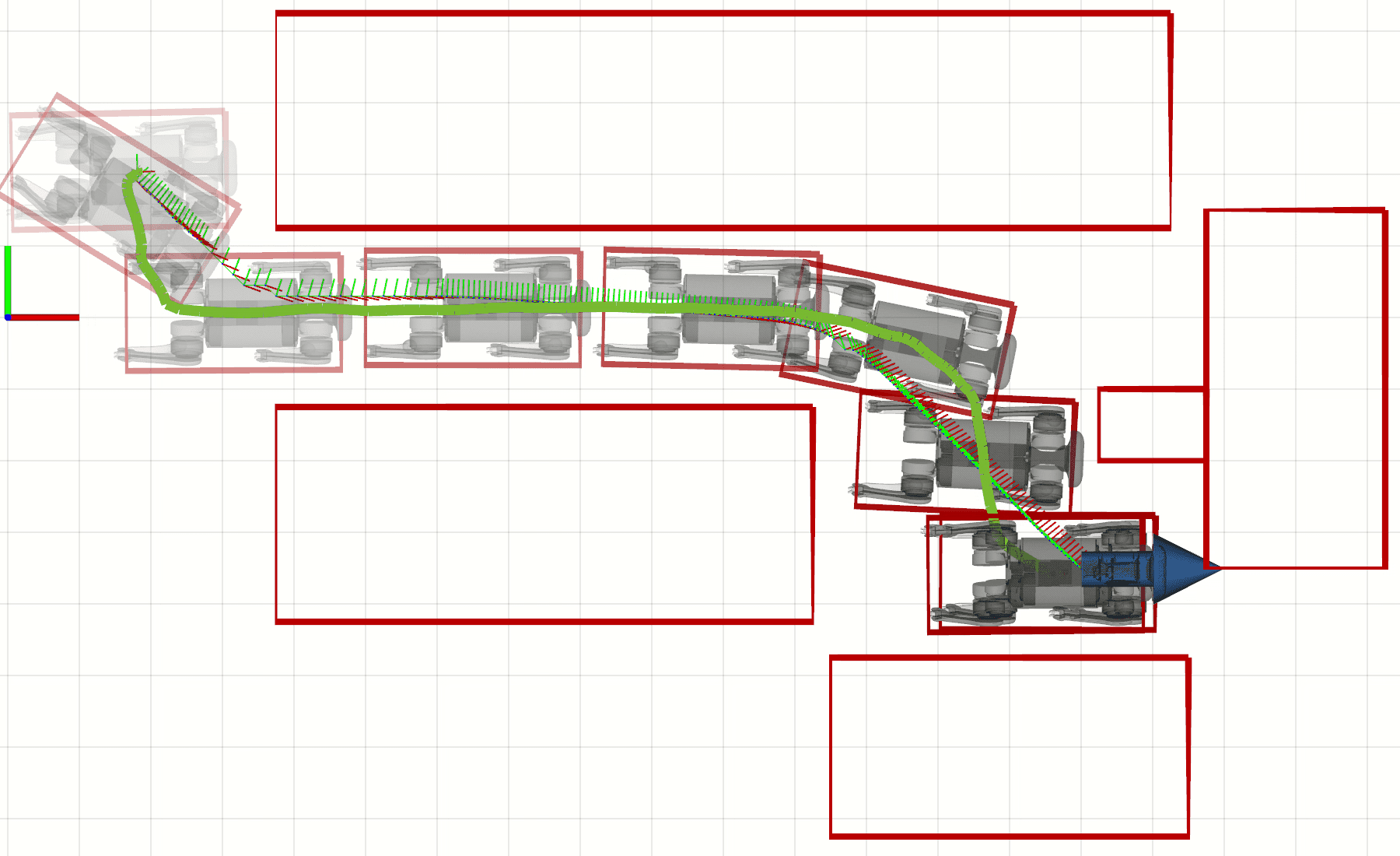}}  node[xshift=0.25\linewidth] at (-0.2\linewidth, -0.15\linewidth) {(d)};
        \draw[xshift=-0.25\linewidth, -{Stealth[scale=1.0]}] (-0.1\linewidth, 0.1\linewidth) node[above,scale=1.0]{Get stuck!} -- (-0.15\linewidth, 0.06\linewidth);
    \end{tikzpicture}
    }
    \caption{Benchmark with collision avoidance methods in narrow spaces. The proposed method uses exponential DCBF Duality that uses CBF on the polytopes avoidance while the baseline uses Euclidean Distance Exponential DCBF where the robot and obstacles are approximated by spheres. In cluttered environments, the robot can get stuck more often and adopt a very slow speed because of the over-approximation using spheres. For example, in (a)(c), the robot gets into a deadlock in narrow spaces. On the contrary, (b)(d) using the proposed method, the robot can freely travel through the same place at high speed because the duality-based CBF allows us to have a finer description of the shapes of the robot and obstacles.}
    \label{fig:scenairo_details}
    \vspace{-15pt}
\end{figure}

In order to showcase the advantages of the polytopic approximation over commonly-used spherical approximation for obstacle avoidance, we compared the proposed navigation framework with the one using Euclidean Distance Exponential DCBF formulation (Fig.~\ref{fig:simple}b) to navigate narrow spaces. The test is deployed in two maps as shown in \figref{fig:scenairo_details}. In the first map, the robot can easily get stuck in the narrow space (Fig.~\ref{fig:scenairo_details}a) because the over-inflation using spheres makes the free space untraversable. In contrast, the proposed method using exponential DCBF Duality can freely travel through the same place because it can consider finer shapes of the robot and environment by polytopes, as illustrated in Fig.~\ref{fig:scenairo_details}b. To quantitatively evaluate these two methods, we perform $48$ trials with random different robot's initial poses, and target poses on this map, and the average traveling time, number of successful trial completion without collision, and failure rate using these two methods are recorded in Table~\ref{tab:random_scenarios}. Results show that the method using polytopic approximation can reach the goal much faster than the spherical baseline ($8.4$ sec versus $13.5$ sec) while having less chance to collide or get stuck in the environment ($20.8\%$ versus $50\%$). Such advantages are further highlighted in the second map, a more challenging scenario with a long narrow corridor as shown in Fig.~\ref{fig:scenairo_details}c,d. The navigation autonomy using the spherical approximation failed in all $16$ trials because the robot gets stuck to entering the corridor, while the proposed method can freely travel through such a tight space and only results in an $8.3\%$ failure rate. Some of the failures using the proposed method are due to an infeasible global path that does not consider the robot's configuration.

Such a quantitative benchmark highlights the advantages of the proposed method, which results in a safe and fast trajectory for the robot to travel in a tighter space that can be untraversable by commonly-used spherical approximation.

\begin{remark}
We must note that one can potentially use an ellipsoid~\cite{brito2019model} for approximating the shapes of the robot and obstacles instead of multiple spheres. However, using single ellipsoids will result in conservative approximations in one of the minor or major axis directions. Super-ellipses~\cite{menon2017trajectory} can also be used for tighter approximations, however, these use fractional powers of distances and could lead to non-smooth changes in the control input and are also sensitive to numerical tolerances.
\end{remark}

\begin{table}[t]
\centering
\caption{The benchmark of the performance using different methods to travel the maps shown in Fig.~\ref{fig:scenairo_details}.}
\label{tab:random_scenarios}

\begin{tabular}{cccc}
\hline
\multicolumn{1}{c|}{\textbf{Method}}             & \textbf{Time (s)}     & \textbf{Completion}   & \textbf{Fail Rate (\%)}  \\ \hline
\multicolumn{4}{c}{\textbf{Map 1}, 48 random initial and target poses}                                                          \\ \hline
\multicolumn{1}{c|}{Duality (Ours)}            & \textbf{8.4} & \textbf{38} & \textbf{20.8\%} \\
\multicolumn{1}{c|}{Euclidean Distance} & 13.5         & 24            & 50.0\% \\ \hline
\multicolumn{4}{c}{\textbf{Map 2}, 16 random initial and target poses}                                                          \\ \hline
\multicolumn{1}{c|}{Duality (Ours)}            & \textbf{9.1} & \textbf{15}   & \textbf{8.3\%}  \\
\multicolumn{1}{c|}{Euclidean Distance} & N/A          & 0            & 100.0\%         \\ \hline
\end{tabular}

\end{table}

\begin{figure}[t]
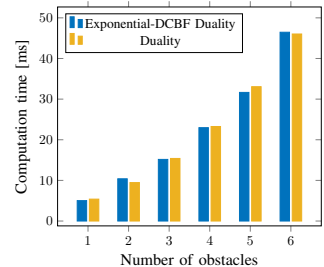

    \centering
    \resizebox{0.5\linewidth}{!}{
    \plotcomputationtime
    }
    \caption{Computation time of NMPC with the different number of obstacles using the exponential DCBF duality (proposed) and the one without CBF (Duality) for obstacle avoidance. The test is deployed on the robot's onboard computer with a regular CPU. The number of vertices of each obstacle is $15$. Using CBFs results in negligible additional computing time of 1.55 ms on average, but it brings clear advantages by allowing the robot to have a smoother trajectory shown in Fig.~\ref{fig:simple}.}
    \label{fig:comp_time}
    \vspace{-15pt}
\end{figure}

\subsection{Computing Speed}
Having seen the advantages of CBFs (Sec.~\ref{subsec:single_sim}) and duality-based formulation (Sec.~\ref{subsec:narrow_sim}), we now investigate the cost of using CBFs: the computing speed. Compared to one simple state constraint, the duality-based CBFs formulated in \eqref{eq:exp_dcbf_duality} introduce additional constraints, which may slow down the solving time critical for online deployment. We measured the computation time of solving NMPC using the proposed method (exponential DCBF Duality) with Duality constraint (no CBF) on the robot onboard computer with Intel Core i7-1165G7 CPU. In \figref{fig:comp_time}, we record the solving time with different numbers of obstacles, each having $15$ vertices. According to \figref{fig:comp_time}, the NMPC computation time increases with the number of obstacles and with the use of CBFs.  However, the usage of exponential DCBF in \eqref{eq:exp_dcbf_duality} only requires an average of $1.55$~ms more solving time than the one without CBF, which is negligible. This study rules out the concern of the potential drawback of CBFs.

\textit{Summary of the Ablation Study:} After the above-mentioned ablation study, we can summarize three points regarding our obstacle avoidance formulation: (i) CBFs makes the robot trajectory much smoother (Fig.~\ref{fig:simple}), (ii) the polytopic approximation by duality-based optimization enables the robot to travel through a tighter space with a faster speed (Fig.~\ref{fig:simple}, Fig.~\ref{fig:scenairo_details}), and (iii) using CBFs results in an insignificant increase in computing time (Fig.~\ref{fig:comp_time}) which enables online deployments on robots. These lead to an exponential DCBF Duality formulation which is the proposed method.

\subsection{Experimental Setup}
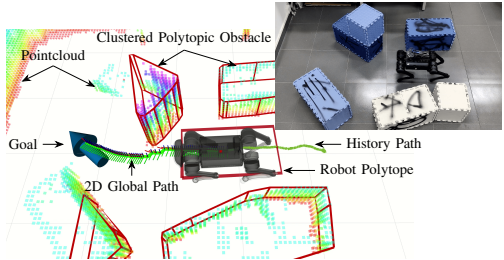
\begin{figure}[!t]
    \centering
    \resizebox{0.8\linewidth}{!}{
    \begin{tikzpicture}
        \node at (0,0) {\includegraphics[width=0.9\linewidth]{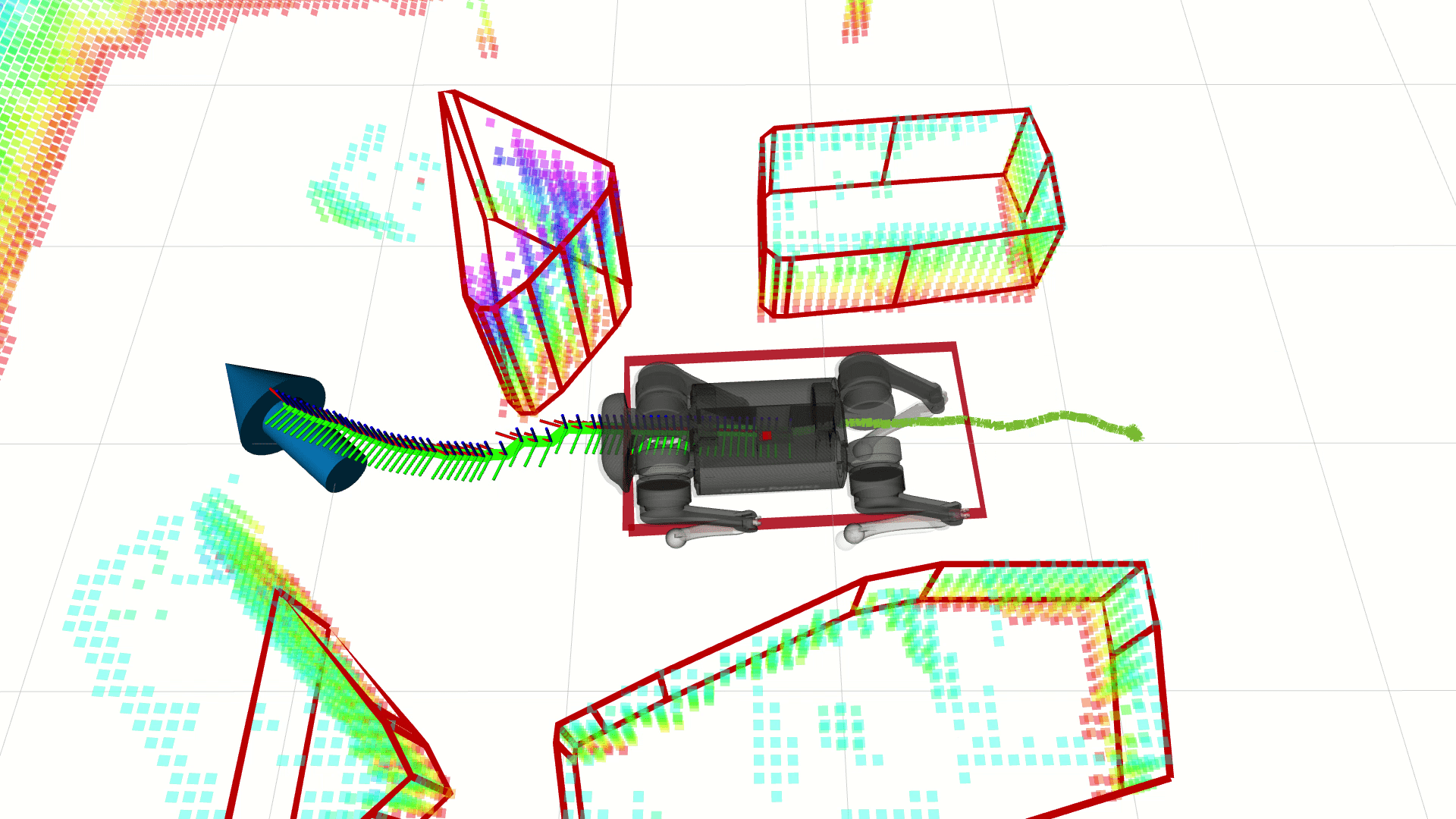}};
        \node at (3.4,1.5) {\includegraphics[width=0.5\linewidth]{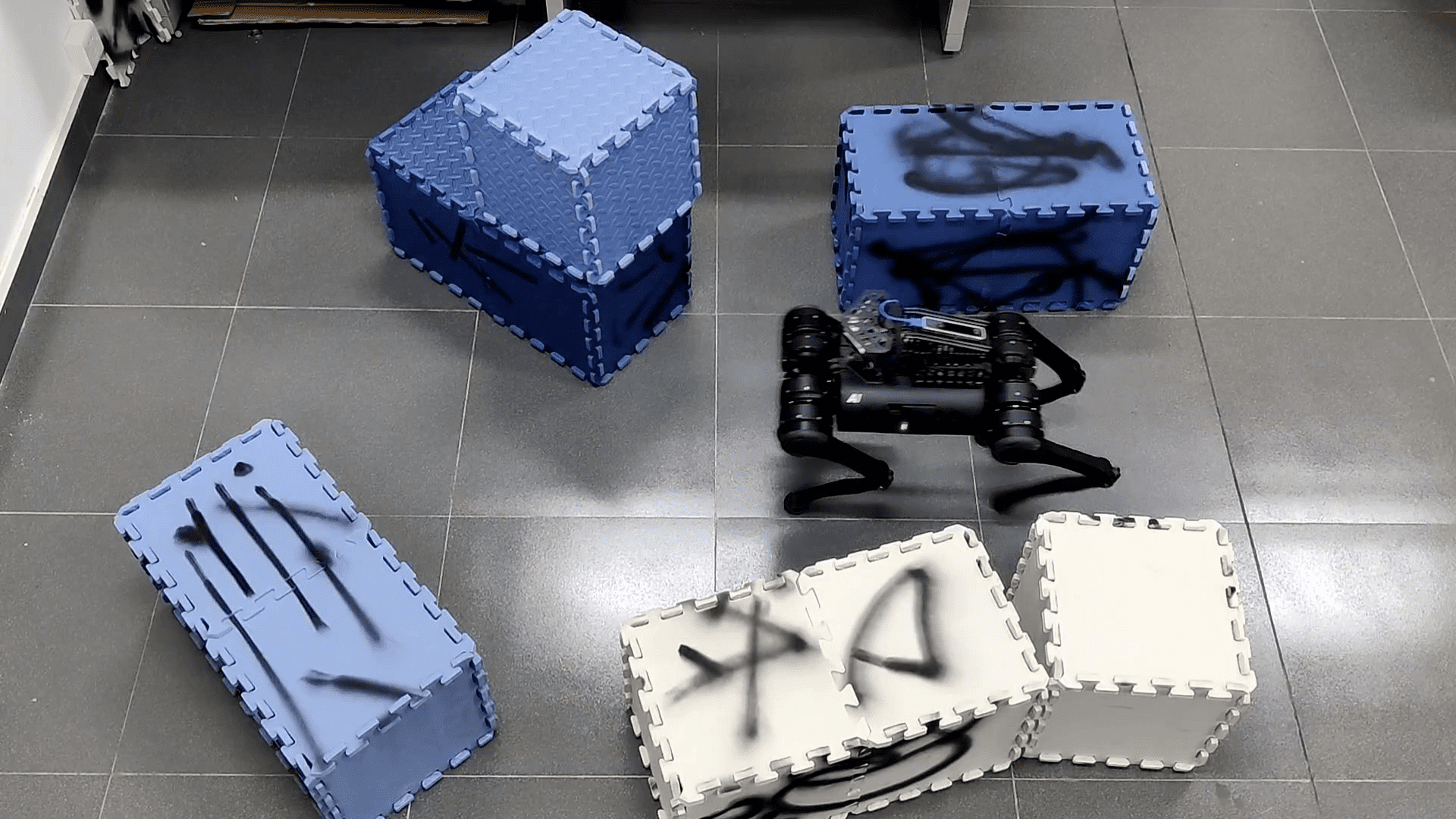}};
        \draw[-{Stealth[scale=1.0]}] (2.5,0)  node[right,scale=0.8]{History Path} -- (2.0,0) ;
        \draw[-{Stealth[scale=1.0]}] (-1.5,-0.6)  node[below,scale=0.8]{2D Global Path} -- (-1.5,-0.3) ;
        \draw[-{Stealth[scale=1.0]}] (-3.2,0)  node[left,scale=0.8]{Goal} -- (-2.8,0) ;
        \draw[-{Stealth[scale=1.0]}] (-0.5,1.8)  node[above,scale=0.8]{Clustered Polytopic Obstacle} -- (0.5,1.5) ;
        \draw[-{Stealth[scale=1.0]}] (-0.5,1.8) -- (-1,1.5);
        \draw[-{Stealth[scale=1.0]}] (2,-0.5)  node[right,scale=0.8]{Robot Polytope} -- (1.4,-0.5) ;
        \draw[-{Stealth[scale=1.0]}] (-3,1.5) node[above,scale=0.8]{Pointcloud} -- (-3.6,1.2);
        \draw[-{Stealth[scale=1.0]}] (-3,1.5) -- (-2.2,1.1);
    \end{tikzpicture}
    }
    \caption{Visualization of the planning data for a quadruped using the proposed framework, where data is taken from an experiment in the real world. The user-commanded 2D goal is drawn as a blue arrow; the path found by the global planner is shown as a chain of axes; the green curve represents the robot torso's history path; polytopes of the robot and obstacles are marked as the red bounding boxes.}
    \vspace{-15pt}
    \label{fig:rviz_label}
\end{figure}

We now deploy the entire pipeline shown in Fig.~\ref{fig:system_dia} on the hardware of A1. The hardware and simulation interface, state estimation, and WBC are open-sourced.
The parameters used in the framework shown in Fig.~\ref{fig:system_dia} are explained as follows. 
The resolution of Octomap is $\SI{0.025}{\m}$, and it generates a projected occupancy grid map with the same resolution.
The global planner re-plans with the linear tolerance of $\SI{0.1}{\m}$.
The timespan for NMPC in Sec.~\ref{sec:formulation} is set to $T = \SI{1.0}{\s}$ is used with a nominal time adaptive discretization of $\delta t \approx \SI{0.015}{\s}$. We use a trotting gait with a $\SI{0.5}{\s}$ gait cycle.
The polytopic obstacles used for exponential DCBF duality constraints are chosen as the closest four obstacles to the robot (within a $\SI{1}{\m}$ box), and the maximum number of vertices of each polytope is set to $15$.
The robot is represented by a rectangle with the $\SI{0.32}{m}$ width and $\SI{0.6}{\m}$ length.
Margins are set to $\alpha =\SI{0.03}{\m} $, $\beta = \SI{0.06}{\m}$ and decay rate $\gamma(\x) = 0.98$ is used in the exponential DCBF constraints. The robot's desired linear velocity is set to $\SI{0.5}{\m\per\s}$.

\subsection{Navigation through Narrow Environments}

\begin{figure}[!t]
    \centering
    \resizebox{\linewidth}{!}{
    \begin{tikzpicture}
    \node[] at (0,0) {\includegraphics[width=\linewidth]{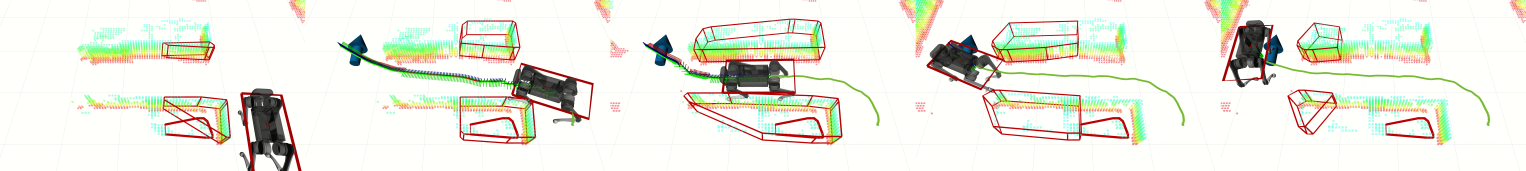}};
    \node[] at (0,-0.11\linewidth) {\includegraphics[width=\linewidth]{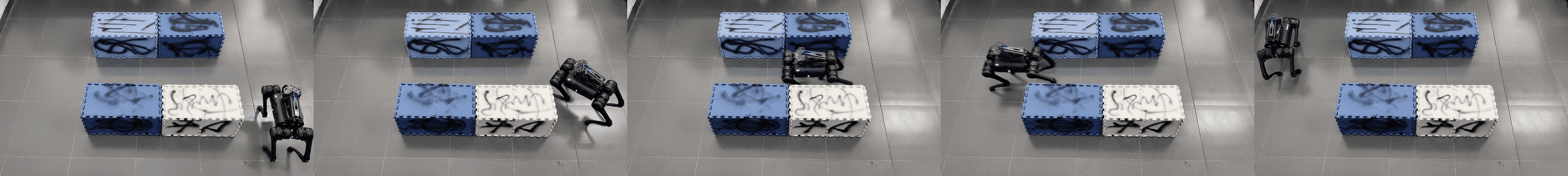}};
    \node[] at (0,-0.22\linewidth) {\includegraphics[width=\linewidth]{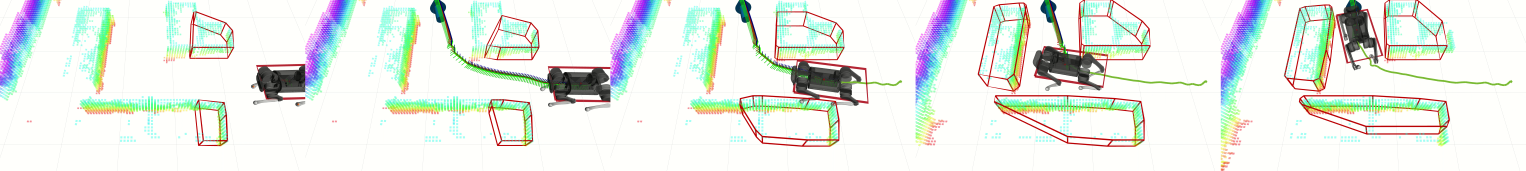}};
    \node[] at (0,-0.33\linewidth) {\includegraphics[width=\linewidth]{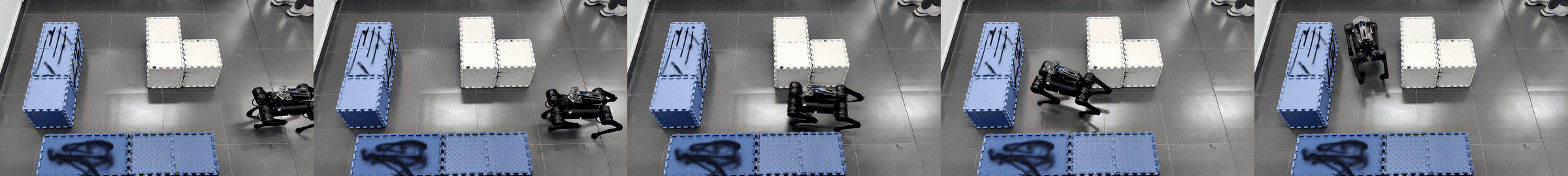}};
    \node[] at (0,-0.44\linewidth) {\includegraphics[width=\linewidth]{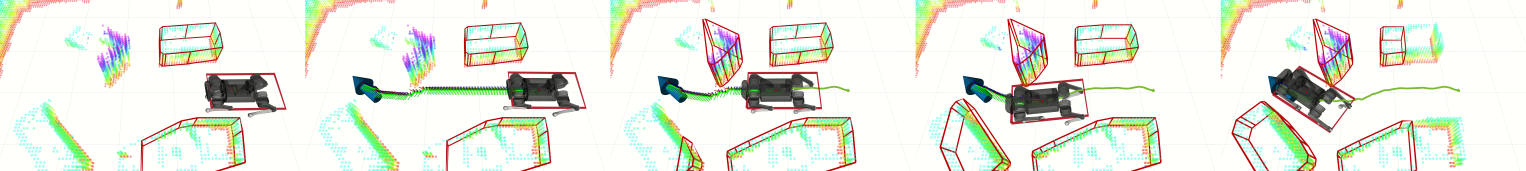}};
    \node[] at (0,-0.55\linewidth) {\includegraphics[width=\linewidth]{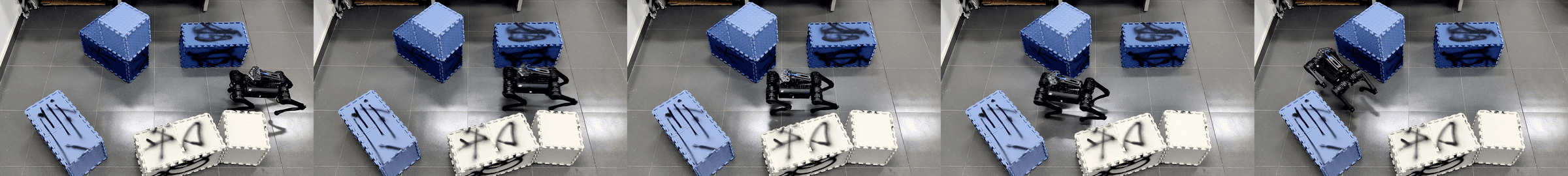}};
    \node[] at (0,-0.66\linewidth) {\includegraphics[width=\linewidth]{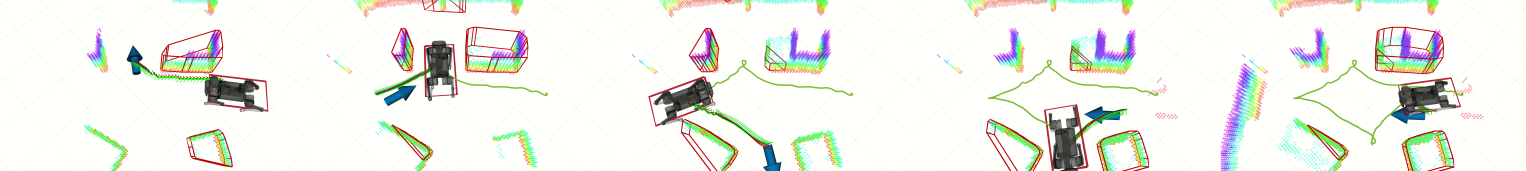}};
    \node[] at (0,-0.77\linewidth) {\includegraphics[width=\linewidth]{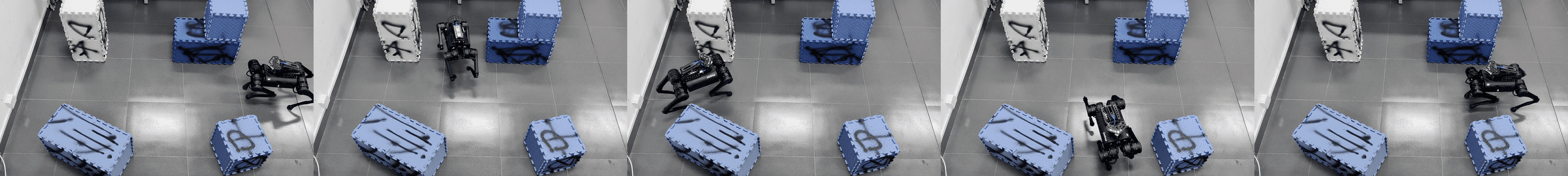}};

    \draw[dashed] (-0.53\linewidth,-0.165\linewidth) -- +(1.03\linewidth,0);
    \draw[dashed] (-0.53\linewidth,-0.385\linewidth) -- +(1.03\linewidth,0);
    \draw[dashed] (-0.53\linewidth,-0.605\linewidth) -- +(1.03\linewidth,0);

    \node[scale=0.7,right,rotate=90] at (-0.52\linewidth,-0.11\linewidth) {Straight};
    \node[scale=0.7,right,rotate=90] at (-0.52\linewidth,-0.33\linewidth) {L-shape};
    \node[scale=0.7,right,rotate=90] at (-0.52\linewidth,-0.55\linewidth) {V-shape};
    \node[scale=0.7,right,rotate=90] at (-0.52\linewidth,-0.77\linewidth) {Random};

    \end{tikzpicture}
    }
    \caption{Planning data visualization from experiments in four scenarios. For each scenario, snapshots of the robot's motion in time are shown. The obstacles are represented by polytopes (a red bounding box) after clustering the pointcloud, and a rectangle bounds the robot. Using the proposed method, the robot can safely and swiftly travel through all of these narrow spaces.}
    \label{fig:exp_rviz}
    \vspace{-15pt}
\end{figure}
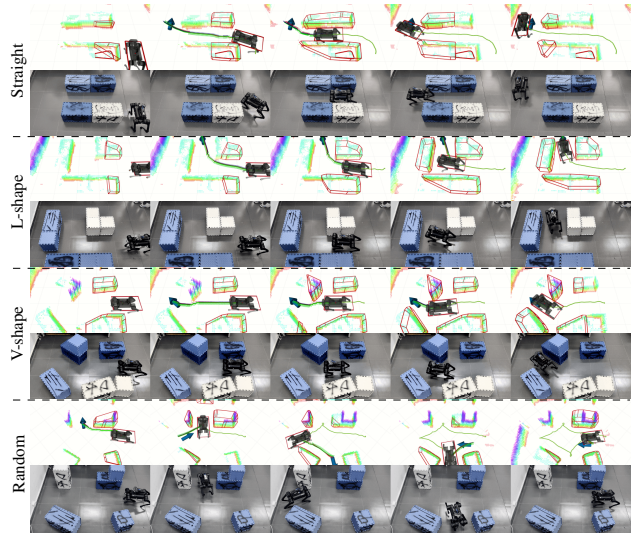

We carried out four experiments in the real world as presented in Fig.~\ref{fig:exp_rviz} and consist of navigating in 1) \textbf{Straight corridor} with two obstacles, a $\SI{0.5}{\m}$ minimum clearance, and $\SI{2.5}{\m}$-long path, 2) \textbf{L-shape corridor} with three obstacles, a $\SI{0.6}{\m}$ minimum clearance, and a path length of about $\SI{2.5}{\m}$.
3) \textbf{V-shape corridor} with four obstacles and a path length about $\SI{2.0}{\m}$, and 4) \textbf{Random obstacles} with four different obstacles and a path length about $\SI{5.5}{\m}$. 

During experiments, the robot was commanded to go through these narrow corridors respectively and return back; we also commanded the robot randomly in the Random corridor.
The experiments are recorded in the accompanying video, and the visualization sequences of environments and states generated by mentioned experiment data are shown in \figref{fig:exp_rviz}, and the visual representations are detailed in \figref{fig:rviz_label}.

As demonstrated in Fig.\ref{fig:exp_rviz}, in the straight corridor trial, the robot can slow down and turn its heading $90$ degrees to fit itself into the corridor. After the robot enters the corridor, it speeds up to the desired speed while avoiding collision with the walls on both sides. 
In the L-shape and V-shape corridors, the robot decelerates at the corners to turn its body slowly without colliding with the cluttered obstacles nearby. 
In the random obstacles trial, the robot showcases the capacity to smoothly navigate in the free space formulated by the random obstacles, even fitting into the gap between adjacent obstacles.
Throughout these experiments, the minimum distance between the robot and obstacles is typically reached when the robot is turning or trying to squeeze between two obstacles.
However, the robot never collides with the obstacle throughout all experiment trials.

\label{sec:results}

\section{Conclusion and Future Works}
In this paper, we proposed a Nonlinear MPC framework based on exponential DCBF duality for safety-critical locomotion control on quadrupedal robots. 
The proposed framework enabled the quadrupedal robot to safely and smoothly walk in narrow spaces by considering the shapes of the robot and the obstacles as polytopes. 
By extensive ablation study, we showed that the introduction of polytopic approximation allows the robot to travel through tighter spaces, and the CBF results in a smoother robot trajectory with an insignificant increase in computing time. This highlights the advantages of the proposed method for navigation and autonomy of legged robots.
We validated our approach on a quadrupedal robot hardware, A1, in various obstacle-laden environments, and the robot shows the ability to maneuver swiftly through these cluttered environments.
However, we have only considered the navigation problem in 2D space. Future work could include implementing 3D obstacle avoidance with polytopes in a tighter space. Moreover, to robustify the proposed method, control errors can be considered like~\cite{kousik2017safe}, and the DCBF Duality constraints can also be added in WBC.

\label{sec:conclusion}

\section*{Acknowledgements} 
The authors thank Xingxing Wang and Unitree Robotics for lending the A1 for experiments and GDUT DynamicX robot team members for their generous help in experiments.

{
\normalem
\bibliographystyle{IEEEtran}
\bibliography{bib/bibliography}

\begin{thebibliography}{10}
\providecommand{\url}[1]{#1}
\csname url@samestyle\endcsname
\providecommand{\newblock}{\relax}
\providecommand{\bibinfo}[2]{#2}
\providecommand{\BIBentrySTDinterwordspacing}{\spaceskip=0pt\relax}
\providecommand{\BIBentryALTinterwordstretchfactor}{4}
\providecommand{\BIBentryALTinterwordspacing}{\spaceskip=\fontdimen2\font plus
\BIBentryALTinterwordstretchfactor\fontdimen3\font minus
  \fontdimen4\font\relax}
\providecommand{\BIBforeignlanguage}[2]{{%
\expandafter\ifx\csname l@#1\endcsname\relax
\typeout{** WARNING: IEEEtran.bst: No hyphenation pattern has been}%
\typeout{** loaded for the language `#1'. Using the pattern for}%
\typeout{** the default language instead.}%
\else
\language=\csname l@#1\endcsname
\fi
#2}}
\providecommand{\BIBdecl}{\relax}
\BIBdecl

\bibitem{tranzatto2022cerberus}
M.~Tranzatto, F.~Mascarich, L.~Bernreiter, C.~Godinho, M.~Camurri, S.~Khattak,
  T.~Dang, V.~Reijgwart, J.~Loeje, D.~Wisth \emph{et~al.}, ``Cerberus:
  Autonomous legged and aerial robotic exploration in the tunnel and urban
  circuits of the darpa subterranean challenge,'' \emph{arXiv preprint
  arXiv:2201.07067}, 2022.

\bibitem{buchanan2021perceptive}
R.~Buchanan, L.~Wellhausen, M.~Bjelonic, T.~Bandyopadhyay, N.~Kottege, and
  M.~Hutter, ``Perceptive whole-body planning for multilegged robots in
  confined spaces,'' \emph{Journal of Field Robotics}, 2021.

\bibitem{huang2021efficient}
J.-K. Huang and J.~W. Grizzle, ``Efficient anytime clf reactive planning system
  for a bipedal robot on undulating terrain,'' \emph{arXiv preprint
  arXiv:2108.06699}, 2021.

\bibitem{tonneau2018efficient}
S.~Tonneau, A.~Del~Prete, J.~Pettr{\'e}, C.~Park, D.~Manocha, and N.~Mansard,
  ``An efficient acyclic contact planner for multiped robots,'' \emph{Trans.
  Robot.}, 2018.

\bibitem{grandia2021multi}
R.~Grandia, A.~J. Taylor, A.~D. Ames, and M.~Hutter, ``Multi-layered safety for
  legged robots via control barrier functions and model predictive control,''
  in \emph{Proc. Int. Conf. Robot. Automat.}, 2021.

\bibitem{teng2021toward}
S.~Teng, Y.~Gong, J.~W. Grizzle, and M.~Ghaffari, ``Toward safety-aware
  informative motion planning for legged robots,'' \emph{arXiv preprint
  arXiv:2103.14252}, 2021.

\bibitem{agrawal2017discrete}
A.~Agrawal and K.~Sreenath, ``Discrete control barrier functions for
  safety-critical control of discrete systems with application to bipedal robot
  navigation.'' in \emph{Robotics: Science and Systems}, 2017.

\bibitem{schulman2014motion}
J.~Schulman, Y.~Duan, J.~Ho, A.~Lee, I.~Awwal, H.~Bradlow, J.~Pan, S.~Patil,
  K.~Goldberg, and P.~Abbeel, ``Motion planning with sequential convex
  optimization and convex collision checking,'' \emph{Int. J. Robot. Res.},
  2014.

\bibitem{kumagai2018efficient}
I.~Kumagai, M.~Morisawa, S.~Nakaoka, and F.~Kanehiro, ``Efficient locomotion
  planning for a humanoid robot with whole-body collision avoidance guided by
  footsteps and centroidal sway motion,'' in \emph{Proc. Int. Conf. Human.
  Robots}, 2018.

\bibitem{oleynikova2017voxblox}
H.~Oleynikova, Z.~Taylor, M.~Fehr, R.~Siegwart, and J.~Nieto, ``Voxblox:
  Incremental 3d euclidean signed distance fields for on-board mav planning,''
  in \emph{Proc. Int. Conf. Intell. Robots Syst.}, 2017.

\bibitem{dudzik2020robust}
T.~Dudzik, M.~Chignoli, G.~Bledt, B.~Lim, A.~Miller, D.~Kim, and S.~Kim,
  ``Robust autonomous navigation of a small-scale quadruped robot in real-world
  environments,'' in \emph{Proc. Int. Conf. Intell. Robots Syst.}, 2020.

\bibitem{kim2020vision}
D.~Kim, D.~Carballo, J.~Di~Carlo, B.~Katz, G.~Bledt, B.~Lim, and S.~Kim,
  ``Vision aided dynamic exploration of unstructured terrain with a small-scale
  quadruped robot,'' in \emph{Proc. Int. Conf. Robot. Automat.}, 2020.

\bibitem{gaertner2021collision}
M.~Gaertner, M.~Bjelonic, F.~Farshidian, and M.~Hutter, ``Collision-free mpc
  for legged robots in static and dynamic scenes,'' in \emph{Proc. Int. Conf.
  Robot. Automat.}, 2021.

\bibitem{chiu2022collision}
J.-R. Chiu, J.-P. Sleiman, M.~Mittal, F.~Farshidian, and M.~Hutter, ``A
  collision-free mpc for whole-body dynamic locomotion and manipulation,'' in
  \emph{Proc. Int. Conf. Robot. Automat.}, 2022.

\bibitem{li2023autonomous}
Z.~Li, J.~Zeng, S.~Chen, and K.~Sreenath, ``Autonomous navigation of
  underactuated bipedal robots in height-constrained environments,'' \emph{The
  International Journal of Robotics Research}, 2023.

\bibitem{xiao2021robotic}
A.~Xiao, W.~Tong, L.~Yang, J.~Zeng, Z.~Li, and K.~Sreenath, ``Robotic guide
  dog: Leading a human with leash-guided hybrid physical interaction,'' in
  \emph{Proc. Int. Conf. Robot. Automat.}, 2021.

\bibitem{ames2019control}
A.~D. Ames, S.~Coogan, M.~Egerstedt, G.~Notomista, K.~Sreenath, and P.~Tabuada,
  ``Control barrier functions: Theory and applications,'' in \emph{European
  Control Conference}, 2019, pp. 3420--3431.

\bibitem{li2022bridging}
Z.~Li, J.~Zeng, A.~Thirugnanam, and K.~Sreenath, ``{Bridging Model-based Safety
  and Model-free Reinforcement Learning through System Identification of Low
  Dimensional Linear Models},'' in \emph{Robotics: Science and Systems}, 2022.

\bibitem{narkhede2022sequential}
K.~S. Narkhede, A.~M. Kulkarni, D.~A. Thanki, and I.~Poulakakis, ``A sequential
  mpc approach to reactive planning for bipedal robots using safe corridors in
  highly cluttered environments,'' \emph{Robot. Automat. Lett.}, 2022.

\bibitem{zhang2021optimization}
X.~Zhang, A.~Liniger, and F.~Borrelli, ``Optimization-based collision
  avoidance,'' \emph{Trans. Control Syst. Tech.}, 2021.

\bibitem{grossmann2002review}
I.~E. Grossmann, ``Review of nonlinear mixed-integer and disjunctive
  programming techniques,'' \emph{Optimization and engineering}, 2002.

\bibitem{gilroy2021autonomous}
S.~Gilroy, D.~Lau, L.~Yang, E.~Izaguirre, K.~Biermayer, A.~Xiao, M.~Sun,
  A.~Agrawal, J.~Zeng, Z.~Li \emph{et~al.}, ``Autonomous navigation for
  quadrupedal robots with optimized jumping through constrained obstacles,'' in
  \emph{Proc. Int. Conf. Automat. Sci. Eng.}, 2021.

\bibitem{yang2022collaborative}
C.~Yang, G.~N. Sue, Z.~Li, L.~Yang, H.~Shen, Y.~Chi, A.~Rai, J.~Zeng, and
  K.~Sreenath, ``Collaborative navigation and manipulation of a cable-towed
  load by multiple quadrupedal robots,'' \emph{Robot. Automat. Lett.}, 2022.

\bibitem{zeng2021safety}
J.~Zeng, B.~Zhang, and K.~Sreenath, ``Safety-critical model predictive control
  with discrete-time control barrier function,'' in \emph{American Control
  Conference}, 2021.

\bibitem{zeng2021enhancing}
J.~Zeng, Z.~Li, and K.~Sreenath, ``Enhancing feasibility and safety of
  nonlinear model predictive control with discrete-time control barrier
  functions,'' in \emph{Conference on Decision and Control}, 2021.

\bibitem{thirugnanam2022safetycritical}
A.~Thirugnanam, J.~Zeng, and K.~Sreenath, ``Safety-critical control and
  planning for obstacle avoidance between polytopes with control barrier
  functions,'' in \emph{Proc. Int. Conf. Robot. Automat.}, 2022.

\bibitem{hornung2013octomap}
A.~Hornung, K.~M. Wurm, M.~Bennewitz, C.~Stachniss, and W.~Burgard, ``Octomap:
  An efficient probabilistic 3d mapping framework based on octrees,''
  \emph{Autonomous robots}, 2013.

\bibitem{barber1996quickhull}
C.~B. Barber, D.~P. Dobkin, and H.~Huhdanpaa, ``The quickhull algorithm for
  convex hulls,'' \emph{ACM Transactions on Mathematical Software (TOMS)},
  vol.~22, no.~4, pp. 469--483, 1996.

\bibitem{orin2013centroidal}
D.~E. Orin, A.~Goswami, and S.-H. Lee, ``Centroidal dynamics of a humanoid
  robot,'' \emph{Autonomous robots}, pp. 161--176, 2013.

\bibitem{bellicoso2016perception}
C.~D. Bellicoso, C.~Gehring, J.~Hwangbo, P.~Fankhauser, and M.~Hutter,
  ``Perception-less terrain adaptation through whole body control and
  hierarchical optimization,'' in \emph{International Conference on Humanoid
  Robots}, 2016, pp. 558--564.

\bibitem{sleiman2021unified}
J.-P. Sleiman, F.~Farshidian, M.~V. Minniti, and M.~Hutter, ``A unified mpc
  framework for whole-body dynamic locomotion and manipulation,''
  \emph{Robotics and Automation Letters}, pp. 4688--4695, 2021.

\bibitem{frison2020hpipm}
G.~Frison and M.~Diehl, ``Hpipm: a high-performance quadratic programming
  framework for model predictive control,'' \emph{IFAC-PapersOnLine}, 2020.

\bibitem{ocs2}
F.~Farshidian \emph{et~al.}, ``{OCS2}: An open source library for optimal
  control of switched systems,'' [Online]. Available:
  \url{https://github.com/leggedrobotics/ocs2}.

\bibitem{carpentier2019pinocchio}
J.~Carpentier, G.~Saurel, G.~Buondonno, J.~Mirabel, F.~Lamiraux, O.~Stasse, and
  N.~Mansard, ``The pinocchio c++ library: A fast and flexible implementation
  of rigid body dynamics algorithms and their analytical derivatives,'' in
  \emph{Proc. Int. Sym. Syst. Integrat.}, 2019.

\bibitem{ferreau2014qpoases}
H.~J. Ferreau, C.~Kirches, A.~Potschka, H.~G. Bock, and M.~Diehl, ``qpoases: A
  parametric active-set algorithm for quadratic programming,''
  \emph{Mathematical Programming Computation}, pp. 327--363, 2014.

\bibitem{brito2019model}
B.~Brito, B.~Floor, L.~Ferranti, and J.~Alonso-Mora, ``Model predictive
  contouring control for collision avoidance in unstructured dynamic
  environments,'' \emph{Robot. Automat. Lett.}, 2019.

\bibitem{menon2017trajectory}
M.~S. Menon, V.~Ravi, and A.~Ghosal, ``Trajectory planning and obstacle
  avoidance for hyper-redundant serial robots,'' \emph{J. Mechan. Robot.},
  2017.

\bibitem{kousik2017safe}
S.~Kousik, S.~Vaskov, M.~Johnson-Roberson, and R.~Vasudevan, ``Safe trajectory
  synthesis for autonomous driving in unforeseen environments,'' in
  \emph{Dynamic systems and control conference}, 2017.

\end{thebibliography}
}

\end{document}